%% file: iclr2026_conference.tex
\documentclass{article} % For LaTeX2e
\usepackage{iclr2026_conference,times}

\input{math_commands.tex}

\usepackage{hyperref}
\usepackage{url}
\usepackage{wrapfig}
\usepackage{graphicx}
\usepackage{xcolor}
\usepackage{duckuments}
\usepackage{booktabs}

\usepackage{listings} % for code blocks
\definecolor{codegreen}{rgb}{0,0.6,0}
\definecolor{codeblue}{rgb}{0.21,0.46,0.66}
\definecolor{codegray}{rgb}{0.5,0.5,0.5}
\definecolor{codepurple}{rgb}{0.58,0,0.82}
\definecolor{backcolour}{rgb}{0.95,0.95,0.92}
\definecolor{gabecheck}{rgb}{0.58,0,0.82}

\lstdefinestyle{mystyle}{
    backgroundcolor=\color{backcolour},
    basicstyle=\ttfamily\scriptsize, % uniform font + size
    commentstyle=,
    keywordstyle=,
    stringstyle=,
    identifierstyle=,
    numberstyle=\scriptsize, % same size for line numbers
    breakatwhitespace=false,
    breaklines=true,
    captionpos=b,
    keepspaces=true,
    numbers=left,
    numbersep=5pt,
    showspaces=false,
    showstringspaces=false,
    showtabs=false,
    tabsize=4
}

\usepackage[ruled,linesnumberedhidden]{algorithm2e}

\title{Programmatic Representation Learning with Language Models}

\author{Gabriel Poesia$^*$ \\
Kempner Institute at Harvard University \\
\texttt{gabriel\_poesia@fas.harvard.edu}
\And
Georgia Gabriela Sampaio\thanks{Authors contributed equally. Code available at \url{https://github.com/gpoesia/leapr/}} \\
Stanford University \\
\texttt{gsamp@stanford.edu}
}

\iclrfinalcopy % Uncomment for camera-ready version, but NOT for submission.

\begin{document}

\maketitle

\begin{abstract}

Classical models for supervised machine learning, such as decision trees, are efficient and interpretable predictors, but their quality is highly dependent on the particular choice of input features. Although neural networks can learn useful representations directly from raw data (e.g., images or text), this comes at the expense of interpretability and the need for specialized hardware to run them efficiently. In this paper, we explore a hypothesis class we call \emph{Learned Programmatic Representations} (LeaPR) models, which stack arbitrary features represented as code (functions from data points to scalars) and decision tree predictors. We synthesize feature functions using Large Language Models (LLMs), which have rich prior knowledge in a wide range of domains and a remarkable ability to write code using existing domain-specific libraries. We propose two algorithms to learn LeaPR models from supervised data. First, we design an adaptation of FunSearch to learn \emph{features} rather than directly generate predictors. Then, we develop a novel variant of the classical ID3 algorithm for decision tree learning, where new features are generated on demand when splitting leaf nodes. In experiments from chess position evaluation to image and text classification, our methods learn high-quality, neural network-free predictors often competitive with neural networks. Our work suggests a flexible paradigm for learning interpretable representations end-to-end where features and predictions can be readily inspected and understood.

\end{abstract}

\section{Introduction}

The central problem in supervised machine learning is to find a predictor $h : X \rightarrow Y$ in a hypothesis class $\mathcal{H}$ that minimizes a certain risk function $\mathcal{R}(h)$, such as $0-1$ error in classification or mean-squared error in regression \citep{michalski2013machine}. Classical choices for $\mathcal{H}$ include linear models, decision trees, and ensembles thereof, which are compellingly simple to understand and debug, and are both compute- and data-efficient. However, their effectiveness is highly limited in domains with low-level, high-dimensional inputs, such as images or text. For these domains, high-quality models are often best learned by first constructing a high level \emph{representation} of an input $x \in X$ using a set of features $f_i : X \rightarrow \mathcal{R}$ that yield a higher-level encoding of the input that predictors can then rely on. While this offers great flexibility, in practice the effort and domain expertise required to \emph{engineer} a good set of features for a particular learning task severely limits the quality of models that can be obtained with classical predictors in high-dimensional input domains without extensive human effort \citep{dong2018feature, cheng2023chess}.

A remarkably successful paradigm that avoids the need for hand-designed feature engineering is \emph{deep learning}, where $\mathcal{H}$ is set to a parameterized family of neural networks of a domain-appropriate architecture. The core advantage of deep learning is the ability of gradient-based optimization to automatically learn useful representations from raw data \citep{Bengiochapter2007,damian2022neural}. Indeed, deep neural networks can be seen as computing a set of complex, non-linear neural features, then applying a simple predictor on top (e.g., the last fully-connected layer, corresponding to a linear model). However, despite being highly effective for \emph{prediction}, neural features have several drawbacks. First, deep neural networks are highly data intensive, and their ability to generalize drops drastically when in-domain data are scarce \citep{oodsurvey}. Second, neural features are not easily interpretable: analyses of large-scale neural networks typically only find a fraction of neurons that seem to encode human-aligned concepts \citep{huben2023sparse}. This limits the potential of neural models to provide faithful \emph{explanations} for their predictions (e.g., express \emph{why} a given $x$ is being classified as $y$), which are important when experts rely on learned models to support high-stakes decision-making \citep{doshi2017towards}.

\begin{figure}
    \centering
\includegraphics[width=1\textwidth]{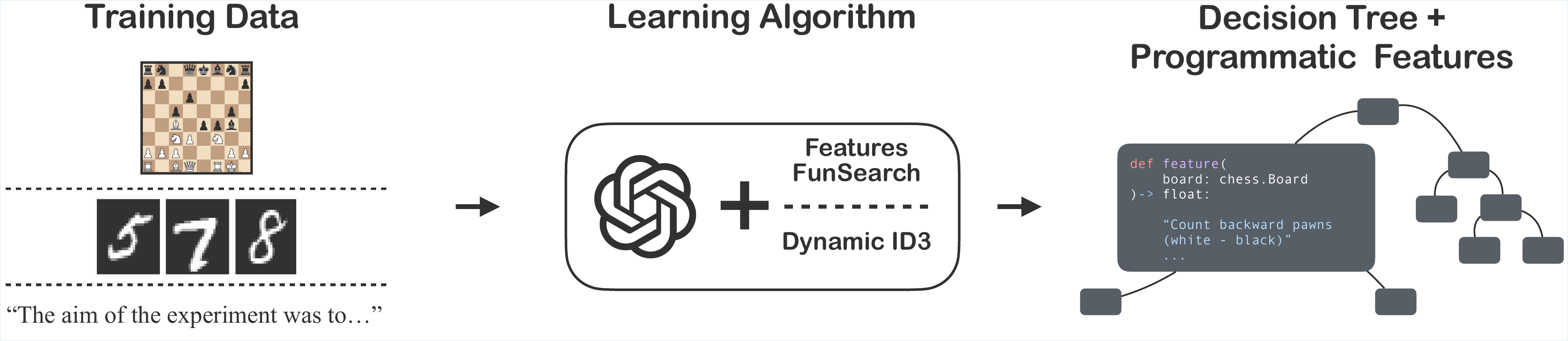}
    \label{fig:overview}
    \vspace{-1em}
    \caption{Learned Programmatic Representation models combine \emph{programmatic features}, synthesized by LLMs as code, and decision tree predictors, yielding interpretable models. We give two algorithms for learning them end-to-end from supervised data in high-dimensional input domains.}
\end{figure}

In this paper, we seek to investigate alternative paradigms for representation learning that, like deep learning, do not require manual feature engineering from users and, like classical methods, yield interpretable, fast, and efficiently-learnable predictors. To that end, we propose learning \emph{LeaPR} (Learned Programmatic Representation) models: a class of predictors with programmatic features paired with decision tree predictors, as illustrated in Figure~\ref{fig:overview}. We leverage Large Language Models' (LLMs) ability to generate code using domain-specific libraries to encode features as arbitrary Python functions from the input domain into the reals: these functions are generated during training with the goal of improving the empirical risk of the current predictor.

For learning from supervised data, we propose two alternative methods. First, we explore a variant of the FunSearch method \citep{romera2024mathematical} called F2 (for \textbf{F}eatures \textbf{F}unSearch). In F2, an LLM iteratively generates functions that compute features from the input domain, which are then evaluated by training a Random Forest \citep{breiman2001random} predictor and extracting importance weights. These scores are used in-context to steer the LLM to generate features with high importance that are thus as predictive as possible. F2 is a \emph{black-box} procedure with respect to the underlying classic predictor, leveraging an off-the-shelf Random Forest training method. We further introduce a \emph{white-box} training procedure for LeaPR models called Dynamic ID3, or D-ID3, inspired by the classical ID3 algorithm for training decision trees. D-ID3 follows an ID3-like training loop where leaf nodes of a growing decision tree are selected and ``split'' by introducing two new leaves and a decision based on the value of a specific feature. While ID3 operates with pre-existing features, D-ID3 queries an LLM to propose new programmatic features on the fly that can help split the node under consideration. D-ID3 gives the LLM the decision path leading to the node, possibly with concrete examples of training samples in that branch, and asks for novel features that are useful for that specific context. The vast prior knowledge of modern LLMs about domain-specific Python libraries makes both approaches highly applicable across diverse, high-dimensional input domains.

We evaluate both methods on two LLMs across three diverse domains: chess position evaluation, image classification (MNIST \citep{lecun1998mnist} and Fashion-MNIST \citep{xiao2017fashion}), and text classification (AI vs. human text classification on the Ghostbuster dataset \citep{verma2024ghostbuster}). Across domains, both LeaPR methods learn high-quality neural network-free representations that extract empirically useful features, often spanning tens of thousands of lines of Python code spread across hundreds of functions, all stitched together by decision trees. Moreover, the learned features tend to be highly intuitive yet specific enough to lead to high-quality predictors. Our main contributions are:

\begin{itemize}
    \item We propose jointly learning \emph{programmatic features}, represented as LLM-generated functions from the input domain to the reals, together with decision tree predictors, thus obtaining fast, interpretable predictors.
    \item We introduce F2 and D-ID3, two algorithms for learning LeaPR models, where features are generated on demand during decision tree training.
    \item We evaluate LeaPR models on three domains: chess positions, images and text, showing comparable accuracy and often favorable data efficiency compared to baseline methods. We analyze the learned features and show how programmatic features can be useful for data exploration and for understanding model failures.
\end{itemize}

\section{Related Work}

\paragraph{Code generation with LLMs}
Our work is enabled by the capability of LLMs to generate code under flexible task specifications, including natural language instructions and examples \citep{chen2021evaluating}. This ability has been explored in a variety of domains, including using code as an intermediate tool for reasoning \cite{chenprogram} and as an interface for agents to interact with an external environment \cite{lv2024codeact}. We exploit code generation for synthesizing \emph{features} from arbitrary input domains, building on LLMs' prior knowledge of a rich set of existing libraries. Feature generation with LLMs has been explored in tabular datasets \citep{ko2025ferg,zhang2024tifg}, whereas we explore domains with complex, low-level input.

\paragraph{Black-box code evolution with LLMs}
LLMs can also be used to \emph{optimize} a black-box objective function \citep{lange2024large}. This approach was pioneered in FunSearch \citep{romera2024mathematical}, the method that inspires our F2 method (Section~\ref{sec:f2}). In FunSearch, the LLM is given the type signature of a function to synthesize, and it outputs candidate functions which are scored with a metric unknown to the model. LLMs can then see past candidate function examples and their scores in following rounds \citep{romera2024mathematical}, and can mutate and evolve previous attempts to propose new, more successful ones. AlphaEvolve \citep{novikov2025alphaevolve} explored this idea with larger models and novel methods to ensure diverse candidates. 
Our work differs from FunSearch and AlphaEvolve in that LeaPR models use LLM-generated code simply to generate \emph{features}, and not the full predictor end-to-end. Using a decision tree to combine LLM-generated features allows our method to scale and simultaneously employ thousands of LLM-generated features at once: for instance, a single model we synthesize in our experiments in chess can use up to $50k$ lines of LLM-generated code to compute features. We believe our work showcases some of the largest fully AI-generated programs written so far as a result of the modularity that our framework provides  (even ignoring the decision trees, which could also be compiled to nested conditionals).

\paragraph{Learning programmatic world models} 
A modular approach for using LLM-synthesized code has been recently employed in PoE-World \citep{piriyakulkij2025poe} for the problem of learning \emph{world models}. In PoE-World, LLMs generate a set of small predictors that individually explain a specific behavior of the environment; a separate model combines these predictors with weights learned by gradient descent. This scales better than monolithic code representations, which WorldCoder introduced \citep{tang2024worldcoder}. Like LeaPR models, PoE-World can generate world models with thousands of lines since LLMs do not have to unilaterally generate all code at once. We propose a similarly modular architecture aimed toward the general problem of supervised learning.

\paragraph{Mechanistic interpretability of neural networks} 
The widespread deployment of deep neural networks strongly motivated the research community to understand the behavior of neural models \citep{mu2021}, often through the lens of \emph{mechanistic interpretability} \cite{conmy2023towards}. One key hypothesis is that neural networks have interpretable ``circuits'' that drive specific behaviors \citep{marks2025} or encode learned knowledge \citep{yao2025}. Prior work has also explored neural architectures that are more amenable to interpretation by construction \citep{doumbouya2025,hewitt2023backpack}, rather than post-hoc. We share this approach of training models that allow for transparent inspection of their behavior, but focusing on a simple composition of programs and decision trees rather than neural networks.

\section{Learning Programmatic Representations}

Classical predictors considered in supervised machine learning, like decision trees, are \emph{intrinsically explainable} in the sense that their structure is simple enough to warrant human inspection: this is highly desirable when humans might want to trust (e.g. in high-stakes decision making) or learn from (e.g., someone learning to play chess) machine learning models. However, this simplicity comes at a steep cost: the performance of these methods is notoriously impacted by the particular choice of features given as their input. 

Consider learning a board evaluation decision tree for the game of chess, for example: a predictor for the likelihood that the player with the white pieces will win in a given board position. During inference, a decision tree proceeds by testing the value of one input variable at a time. If given only the raw information available on the board (e.g. the piece lying on each of the 8x8 squares), each input variable alone carries negligible information about the overall situation in the game. Each leaf node is forced to be invariant to all features that are not tested on the path from the root to that node; thus, if any decision is made before testing \emph{all} 64 squares, there is often a chance that something critical was missed (e.g., a queen on one of the unobserved squares). Thus, this representation is too low-level for decision trees to be useful, even though all that there is to know about the board is derivable from the input. 

On the other hand, if we instead train the model on a \emph{single useful feature} computed from the board, like the \emph{material difference} between players (given by a weighted sum of the pieces of each color still on the board), suddenly a very shallow decision tree can easily learn to make predictions that are significantly better than random, since a large material advantage is highly correlated with one's chance of winning. Adding more informative features will progressively allow a decision tree learner to find better steadily predictors.

In this paper, our starting point is the insight that LLMs have two capabilities that allow them to serve as highly effective \emph{feature engineers} for classical ML models. First, many useful features, such as the material balance in chess described above, can be implemented in a few lines of code, and modern LLMs are capable of flexibly generating code to accomplish a variety of tasks. Second, broad pre-training of LLMs encodes useful prior knowledge about a very wide range of domains \citep{bommasani2021opportunities}, equipping them with strong priors over what kinds of features could be useful for making predictions across these domains. Our main hypothesis is that we might be able to train high-quality classical models by leveraging LLM-generated features at scale. Our goal here is thus to learn a hypothesis class that consists of (a) \emph{programmatic features} and (b) decision tree predictors. We call these \emph{Learned Programmatic Representation} models, or \emph{LeaPR} models. The main challenge we now tackle is how to elicit predictive features from language models.

\subsection{Black-box feature generation}
\label{sec:f2}

Given a supervised dataset $\mathcal{D}$ and any scoring function $S_\mathcal{D}(f)$ that measures the quality of a proposed \emph{feature} $f : X \rightarrow \mathbb{R}$, a simple methodology to obtain increasingly good features is to apply a FunSearch-style procedure, where an LLM is used to propose candidates to try to maximize $S$. Modern LLMs are capable of generating complex functions even for high-dimensional input domains, such as images and text, partly due to their ability to write code that uses existing human-written libraries for each of these domains. Thus, the main component needed for this approach to work is to answer: what makes $f$ a good feature?

% For the purpose of learning a \emph{predictor} mapping $X$ into $Y$, $f(x)$ must have some positive mutual information with the output $y$ (or, more generally, with the distribution over $y$ conditioned on $x$). But classical Shannon-mutual information does not account for computational (or learnability) constraints: for instance, the output of a hash function is entirely determined by its input, yet it is computationally intractable to learn anything about the input by observing its hash. A more useful intuition for our purposes is the concept of \emph{predictive $\mathcal{V}$-information} between $f(x)$ and $y$, which takes into account the additional constraint that we are interested in a predictor from a particular class $\mathcal{V}$ (here, decision trees). Thus, for learning a LeaPR model, a good feature is one that helps a decision tree learning algorithm find a model that achieves a lower risk for the task at hand.

In the standard FunSearch setup \citep{romera2024mathematical}, the scoring function $S$ evaluates candidates independently: proposals are \emph{self-contained solutions} of the task. Our setup here is different: for the purpose of learning a predictor from \emph{all} generated features at once, a new candidate $f_k$ is valuable only to the extent that it contributes predictive information when taking into account the existing feature set $f_{1:k-1}$. Assuming that we keep track of the set of features generated so far and propose new features in a loop, a naïve adaptation of FunSearch would thus score a new feature $f_k$ on its \emph{added} predictive power once it has been proposed. For that, we could train a new predictor using $f_{1:k}$ and compare its risk against the previous predictor trained on $f_{1:k-1}$. However, this runs into the issue that early features receive disproportionately high scores simply because their baselines (initial predictors based on few features) are severely limited. In practice, with this approach, the highest-scoring features are essentially fixed after the first few iterations, which is undesirable: we would like to detect and reward powerful features even if they appear late during training.

To overcome this problem, we \emph{simultaneously score all existing features} $f_{1:k}$ independently of the order in which they were proposed. Given a learned decision tree, prior work has proposed several metrics of \emph{importance} of each input feature (e.g., measuring decrease in ``impurity'' in decision nodes that use a given feature, \citep{breiman2001random}). Importance metrics only depend on the final learned predictor, and decision tree learning methods are order-invariant with respect to input features.

Algorithm~\ref{fig:algo-f2} (Figure~\ref{fig:algo}, left) shows \emph{Features FunSearch} (F2), our representation learning algorithm based on this FunSearch-style approach but with features scored as a set. Specifically, F2 takes a supervised dataset and learns a programmatic representation --- i.e. a set of feature functions $f_i : X \rightarrow \mathbb{R}$, represented as executable code. Like FunSearch, F2 iteratively uses an LLM to make batches of proposals conditioned on a sample of existing features, which are shown to the model along with their assigned scores --- the LLM's task is to propose new features that will be assigned high importance score in a newly trained Random Forest predictor. These scores are a \emph{global estimate} of the predictive power of each feature in a predictor trained with all of them.

\begin{figure}[t]
\begin{minipage}[t]{0.465\textwidth}
    \begin{algorithm}[H]
    \caption{Features FunSearch (F2)\label{fig:algo-f2}}
% 	\LinesNumbered
	{\footnotesize
        \SetKwInOut{Inputs}{Input}
	\SetKwInOut{Output}{Output}
        \Inputs{Language model, $LM$, \\ Supervised dataset $D \in 2^{X \times Y}$}
	    \Output{List of features $F$, where $f_i : X \rightarrow \mathbb{R}$}
    $F \gets [\ ]$\;
    \For {$\mathit{iteration} \in [1, \cdots{}, T]$} {
        $rf \gets \text{TrainRandomForest}(D, F)$ \;
        $imp \gets \text{FeatureImportances}(rf)$ \;
        $top\_k \gets \text{TopKFeatures}(F, imp)$\;
        $r \gets \text{RandomKFeatures}(F\setminus{}top\_k , imp)$\;
        $p \gets \text{ProposeFeatures}(LM, top\_k, r)$\;
        $F.\texttt{extend}(p)$\;
    }
    \Return $F$ \;
   }
\end{algorithm}
\end{minipage}%
\quad
\begin{minipage}[t]{0.5\textwidth}
    \begin{algorithm}[H]
    \caption{Dynamic ID3 (D-ID3)\label{fig:algo-did3}}
% 	\LinesNumbered
	{\footnotesize
        \SetKwInOut{Inputs}{Input}
	    \SetKwInOut{Output}{Output}
        \Inputs{Language model $LM$, \\ Supervised dataset $D \in 2^{X \times Y}$}
        \Output{List of features $F$, where $f_i : X \rightarrow \mathbb{R}$}
    $T \gets \texttt{Leaf}(D_{train})$\;
    \For {$\mathit{iteration} \in [1, \cdots{}, T]$} {
        $\mathit{l} \gets \arg\max_{\texttt{IsLeaf}(n)} \text{TotalError}(T, n.data)$\;
        $p \gets \text{ProposeFeatures}(LM, l.path\_to\_root)$ \;
        $\langle f, t \rangle \gets \arg\min_{f \in F, t} \text{SplitError}(f, t, n.data)$\;
        $l.\text{split}(f, t, \{x \in n.\text{data} | f(x) < t\},$ \\
        $\hspace{4.5em}\{x \in n.\text{data} | f(x) > t\})$ \;
    }
    \Return $\{ n.\text{splitting\_feature} | n \in T \wedge \text{Internal}(n) \}$ \;
    }
    
\end{algorithm}
\end{minipage}
\vspace{-0.6em}
\caption{Two learning algorithms for LeaPR models. F2 (left) uses a FunSearch-style loop that attempts to evolve features that are \emph{globally useful} to train a Random Forest predictor, as estimated by feature importances. D-ID3 (right) runs an ID3-style decision tree training loop and attempts to propose new features that are \emph{locally useful} for splitting specific leaf nodes, attempting to minimize their impurity (e.g., variance in regression, or entropy in classification).}
\label{fig:algo}
\end{figure}

\subsection{Dynamic splitting}

While F2 uses an underlying decision tree learner as a black-box, the insight that LLMs can be used to generate features on demand can serve as the basis for designing the decision tree learner itself. Recall that during inference in a decision tree, we start at the root and repeatedly follow the ``decisions'' associated with each node until we reach a leaf. Each such decision consists of testing the value of a particular feature: if this node splits on feature $f_k$, we compare $f_k(x)$ with a threshold $t$ learned during training. If $f_k(x) < t$, the node recursively returns the prediction made by its left child (or right child if $f_k(x) \geq t$). Leaf nodes return a fixed prediction defined during training, e.g., the most common class label (classification), or average value (regression) for training points that fall on that leaf. For training, classical decision tree learning algorithms (e.g., ID3 or CART \citep{quinlan1986induction}) start with a single node and repeatedly improve the current decision tree predictor by (a) choosing a leaf node and (b) partitioning it into a new decision node with two new leaves as its children. Partitioning searches for a feature and comparison threshold that minimize the ``impurity'' (e.g., variance in the continuous case, or entropy of class labels in classification) of data points falling on both sides. For instance, in classification, the best-case scenario would be to find a partition where all training data points falling on each new leaf belong to the same class.

However, the ability of classical algorithms to find good splits is limited by the predictive power of preexisting dataset features. Here, we revisit this recursive splitting strategy considering that we can attempt to generate new features \emph{on demand} for the purpose of successfully partitioning a particular leaf node. When we decide to split a leaf, we have significant local context aside from global dataset information: in particular, we know the specific path of decisions that leads to that leaf, and we have a corresponding set of training examples, with their labels, falling onto that node. To be \emph{locally useful}, a feature only needs to help distinguish between examples in that set. Indeed, for informing the proposals of potentially useful features, we can even leverage the ability of LLMs to performe inductive reasoning, by presenting actual examples (if possible in text), along with their labels, in the model's context: its objective, then, is to propose a feature that would explain the variation in the labels between those examples and others that reach the same leaf.

Algorithm~\ref{fig:algo-did3}, Dynamic ID3 (D-ID3), realizes this idea. In each iteration, D-ID3 selects the current leaf in the tree that accounts for the largest portion of training error (e.g. number of misclassified training examples). D-ID3 then generates new candidate features with an LLM on the fly to split that particular leaf. In modalities where we can easily represent examples in text, the LLM receives a sample of examples and their labels that fall in this branch (in our experiments, this only excludes image classification, where we only show a sample of image \emph{class labels} in the prompt). D-ID3 considers these features, as well as all candidate features generated for ancestor nodes, and finds the best split for this leaf according to a user-defined impurity metric (all metrics available for classical methods are also possible here). This process repeats for a number of iterations. At the end, like F2, D-ID3 returns a learned \emph{representation}: the set of programmatic features for the input domain that were used in the resulting decision tree. We note several practical considerations for both F2 and D-ID3, as well as other implementation details, in Appendix~\ref{app:practical}. 

\section{Experiments}

We now evaluate programmatic representation models on three tasks with complex input domains where the standard practice is to train neural networks: chess position evaluation (given a chess board, predict the probability that White wins), image classification on MNIST \citep{lecun1998mnist} and Fashion-MNIST \citep{xiao2017fashion}, and text classification (detecting whether a piece of text is human- or AI-generated) on Ghostbuster \citep{verma2024ghostbuster}. In all domains, we compare LeaPR against standard neural network baselines; additionally, in chess and image classification, we also include the Random Forest baseline where we feed a simple ``raw'' encoding of the input (the piece in each square for chess boards, or pixel values for images). We discuss the features our methods learn in each domain, and finally conduct a case study debugging a classifier that has learned to rely on a spurious feature in the Waterbird dataset \citep{sagawa2020waterbirds}.

We run F2 and D-ID3 using two OpenAI models, for a total of 4 LeaPR models per task: GPT 4o-mini \texttt{gpt-4o-
mini-2024-07-21} \citep{hurst2024gpt4o} and GPT 5-mini \texttt{gpt-5-mini-2025-08-07} \citep{openai2025gpt5}. We run both methods so that they output a maximum of 1000 features --- this means using 1000 iterations of D-ID3, and 100 iterations of F2 with a proposal batch size of 10 features in each call. We sometimes end with fewer than 1000 features because we discard features that fail validation (see Section~\ref{app:practical}) Using the features learned by either algorithm, we then train a Random Forest model using the standard Scikit-Learn \citep{scikit-learn} implementation, with 500 trees and a maximum depth of 50.

\begin{wraptable}{O}{0.65\linewidth}
    \centering
    \setlength\tabcolsep{3pt}
    \footnotesize
    \begin{tabular}{l c c c c}

\toprule
% \textbf{LeaPR Algorithm} &  \textbf{Feature Model}
\multicolumn{1}{c}{\textbf{Predictor}}
     &\textbf{Training Size} & \textbf{RMSE} & \textbf{$\rho$} & \textbf{Acc.} \\

        \cmidrule(lr){1-1} \cmidrule(lr){2-2} \cmidrule(lr){3-5}
         \multicolumn{1}{c}{Random policy} & 0 &  & & 11.4\% \\
\multicolumn{1}{c}{Transformer} & $5 \times 10^7$ & .161 & .795 & 30.3\% \\
\multicolumn{1}{c}{Transformer} \citep{ruoss2024amortized} & $5 \times 10^8$ &  &  & 58.5\% \\
\multicolumn{1}{c}{Random Forest (raw board)} & 200k & .248 & .306 & 14.5\% \\
        \cmidrule(lr){1-1} \cmidrule(lr){2-2} \cmidrule(lr){3-5}

        \textbf{LeaPR} F2 + GPT 5-mini & $200k$ & .169 & .762 &  31.4\% \\
         \hspace{3em} F2 + GPT 4o-mini & $200k$ & .163 & .783 &  16.7\% \\
        \hspace{3em} D-ID3 + GPT 5-mini & $200k$ & .160 & .789 & 33.5\% \\
          \hspace{3em} D-ID3 + GPT 4o-mini & $200k$ & .156 & .806 & 17.2\% \\
         %LeaPR + D-ID3 + MCTS (10k) & & & 45\% \\
         
         \bottomrule

    \end{tabular}
    \caption{Performance in state-value prediction models in chess positions from Lichess. We train the 270M-parameter Transformer from \citet{ruoss2024amortized} with up to $50M$ data points, and match their results for their full run on $10x$ more data.}
    \label{tab:chess}
\end{wraptable}

\subsection{Chess position evaluation \label{sec:exp-chess}}

First, we train models on the regression task of state-value prediction in the game of chess: given the board position, predict the win probability for each player. We use a publicly available dataset of games from the Lichess online platform \citep{lichess}, and hold out 1000 random board positions for evaluation. The dataset comes with state values estimated by Stockfish \citep{stockfish2008}, the strongest publicly available chess engine. We use Stockfish's prediction value as the ground truth (Stockfish outputs values in ``centipawns'', which we convert to win percentages using the standard formula used by Lichess and other prior work). To represent and manipulate chess boards, we use the popular \texttt{python-chess} library, with a standard API that facilitates iterating through the board, locating pieces, generating available moves, and testing for various pieces of game state (e.g., a player's turn, whether the current player is in check, etc). Our prompts contain a short listing of the main API classes, methods, and functions available in the library. The models are instructed to generate features that receive an argument of type \texttt{chess.Board} and return a \texttt{float} value. We provide full prompts in Appendix~\ref{app:prompts}.

\textbf{Transformer baseline.} As a neural baseline, we train the 270M parameter Transformer architecture proposed in \citet{ruoss2024amortized} (their largest model) to predict the discretized win-probability for White (128 buckets) given a position encoded in the standard FEN format. \citet{ruoss2024amortized} compared models that predict both state values and state-action values; when controlled for the number of data points, models trained on state-value slightly outperformed when playing games against each other. Their strongest model (trained on 15.3B state-action values) achieved grandmaster-level play.  Due to computational constraints, we reproduce their training of a state-value prediction Transformer run only up to 50M data points (10x less than their total state-value dataset of 500M data points; though we approximately match their number of epochs over the training data at 2.5).

Table~\ref{tab:chess} summarizes the results for this regression task. Here, we show both root mean square error (RMSE) and Pearson correlation ($\rho$) between model predictions and Stockfish's estimate. Our LeaPR models, trained on $200k$ board positions, compare favorably with the Transformer predictor trained on $250x$ more data. In contrast, as expected, Random Forests trained on the raw board struggle. LeaPR models benefit from the significant prior knowledge that LLMs have about useful chess concepts. We see basic features such as one that ``Calculates the total piece value of both sides'' (the model's own function documentation string) proposed and implemented by GPT 4o-mini in 6 lines of code early during training, as well as significantly more complex, specific features such as ``Pawn promotion pressure: sum over pawns of 1/(1+steps$\_$to$\_$promotion) weighted by being passed (white minus black). Encourages advanced, passed pawns.'', implemented by GPT 5-mini late in the D-ID3 run (with 51 lines of code). Generally, D-ID3 features appear to become more specific as training progresses, likely because the LLM is asked to distinguish only a subset of board positions that already share many similarities (due to falling on a specific leaf node), yielding better models than F2 even with the same number of total features.

\begin{wraptable}{O}{0.45\linewidth}
    \centering
    \setlength\tabcolsep{3pt}
    \footnotesize
    \begin{tabular}{l c c c c}
    \toprule
    \multicolumn{1}{c}{\textbf{Predictor}}
     & \textbf{MNIST} & \textbf{Fashion}\\ 
    \cmidrule(lr){1-1} \cmidrule(lr){2-3}
    \multicolumn{1}{c}{ResNet-50}
          & 98.71\% & 89.54\% \\
%                 & ImageNet & 98.28 & 89.51 \\
\multicolumn{1}{c}{EfficientNetV2}
          & 98.8\% & 90.94\% \\
          \multicolumn{1}{c}{Random Forest (raw pixels)}
          & 95.6\% & 88.29\%  \\

\cmidrule(lr){1-1} \cmidrule(lr){2-3}
        % \textit{Algorithm}&\textit{LeaPR Model}& \multicolumn{2}{c}{}\\
        \cmidrule(lr){1-1} \cmidrule(lr){2-3}
          \textbf{LeaPR} F2 + GPT 5 mini & 92.54\% & 85.77\% \\ 
          \hspace{3em}F2 + GPT 4o mini & 89.26\% & 80.26\% \\
          \hspace{3em}D-ID3 + GPT 5 mini & 96.91\% & 88.51\% \\
          \hspace{3em}D-ID3 + GPT 4o mini & 93.71\% &  83.80\% \\
         \bottomrule
    \end{tabular}
    \caption{Top-1 accuracy on image classification on MNIST and Fashion-MNIST.}
    \label{tab:mnist}
\end{wraptable}

We also compare models in terms of Top-1 move accuracy compared to Stockfish at its maximum strength. Since we only estimate state-values, to use our predictors we select the move that leads to the best successor value from the point of view of the current player (i.e., the move leading to the highest or lowest win-probability for White depending on who plays), and measure how often this matches Stockfish's top move. Interestingly, we find that regression performance is not necessarily predictive of move accuracy: LeaPR models trained with GPT 4o-mini are significantly worse when used to select moves. Our best action predictors achieve non-trivial move selection accuracy: whereas random performance for this task is 11.4\%, the D-ID3 model trained with GPT 5-mini predicts the top Stockfish move in 33.5\% of the cases, with the Transformer baseline underperforming at 30.3\%.\citet{ruoss2024amortized} trained this same state-value model with up to 500M data points and managed to achieve a move accuracy of 58.5\%, showing that the Transformer keeps improving for much longer. Their most accurate model for action prediction is trained to directly predict action-values from 15B training data points, achieving an accuracy of 63.5\% and a grandmaster-level ELO rating when playing with humans online. Although there remains a significant gap between their best results achieved with a Transformer and what we demonstrate here, LeaPR models still get surprisingly far in this challenging regression task. If the scalability challenges associated with LeaPR models can be overcome (e.g., we see negligible benefits from training Random Forests beyond $200k$ training data points), we might be able to obtain chess policies that not only play at a high level but can also ``explain'' their moves --- a feature that no existing chess engine possesses.

\subsection{Image classification}
\label{sec:exp-img}

We now evaluate LeaPR models on two image classification datasets: MNIST (handwritten digit classification) from \citet{lecun1998mnist} and Fashion-MNIST (grayscale fashion products classification) from \citet{xiao2017fashion}. We train standard ResNet-50 and EfficientNetV2 baselines to convergence on the same datasets (training details in \ref{app:mnist}). Unlike in Section~\ref{sec:exp-chess}, D-ID3 does not add \emph{images} in the prompt when calling the LLM, but only a textual description of class labels of a set of training examples belonging to the leaf being split (e.g., ``digit 0'' in MNIST, or ``T-shirt'' in Fashion-MNIST). This is a significant limitation for this domain, since the features need to rely solely on the LLM's prior knowledge about what the described objects might look like and hypotheses about how they will be detectable in a small grayscale image. Still, the best LeaPR models in this task achieve comparable accuracy to the neural baselines, even without the ability to construct features by directly observing the training data. Again, especially with D-ID3 we observe specific features that attempt to distinguish between particular classes, such as \emph{``Count of ink "endpoints": ink pixels with only one ink neighbor (8-connected). Loops like 0/8 have few endpoints; open strokes like 5 have endpoints''} being the feature with the best split (thus selected) in a leaf where the majority classes were 8 and 5. GPT 5-mini correctly implemented the above feature in 30 lines of Python using \texttt{numpy}. Though MNIST and Fashion-MNIST generally only serve as ``sanity checks'' for computer vision models (with even Random Forests trained on the raw pixels performing near the neural baselines, given the small image and training set sizes), we believe that this result presents an encouraging signal towards the ability of LeaPR models to achieve comparable accuracy while proposing simple and interpretable programmatic features.

\begin{wraptable}{O}{0.34\linewidth}
    \centering
    \footnotesize
    \setlength\tabcolsep{3pt}
    \begin{tabular}{l c c c c}
    \toprule
    \multicolumn{1}{c}{\textbf{Predictor}} & \textbf{F1} \\ 
    \cmidrule(lr){1-1} \cmidrule(lr){2-2}
    \multicolumn{1}{c}{Perplexity only}
    & 81.5\\
         \multicolumn{1}{c}{GPTZero} & 93.1\\
         \multicolumn{1}{c}{RoBERTa} & 98.1\\
         \multicolumn{1}{c}{Ghostbuster}& 99.0\\
         \cmidrule(lr){1-1} \cmidrule(lr){2-2}

% rounding to the 1st sig fig
\textbf{LeaPR} F2 + GPT 5-mini & 97.7\\ 
         \hspace{3em} F2 + GPT 4o-mini & 98.6\\
         \hspace{3em} D-ID3 + GPT 5-mini & 98.8\\
         \hspace{3em} D-ID3 + GPT 4o-mini & 98.6\\
         \bottomrule
    \end{tabular}
    \caption{F1 score on Ghostbuster: classifying text as human or AI written \citep{verma2024ghostbuster}.}% 
    \label{tab:ghostbuster}
\end{wraptable}
\subsection{Text classification \label{sec:exp-text}}
We now evaluate LeaPR models on a binary text classification task: detecting whether the input was written by an LLM or a human. We use the Ghostbuster dataset \citep{verma2024ghostbuster}, which contains a collection of student essays, creative writing, and news articles written by humans and by ChatGPT and Claude \citep{claude} given the same or similar prompts. In Table~\ref{tab:ghostbuster} we compare LeaPR models with the results for the Ghostbuster model and other neural baselines reported in \citet{verma2024ghostbuster} for the ``in distribution'' setting with all domains combined (we omit DetectGPT, which achieves a low F1 score of 51.6\% due to having been trained to detect another LLM). Here, LeaPR models are the only neural network-free predictors. Still, our models perform competitively when evaluated in F1 score: they outperform all baselines except for Ghostbuster itself, with the best LeaPR model (with features obtained by D-ID3 with GPT 5-Mini) closely matching Ghostbuster (98.8 vs 99.0 in F1 score).

\begin{figure}
    \vspace{-1em}
    \centering
    \includegraphics[width=\linewidth, keepaspectratio]{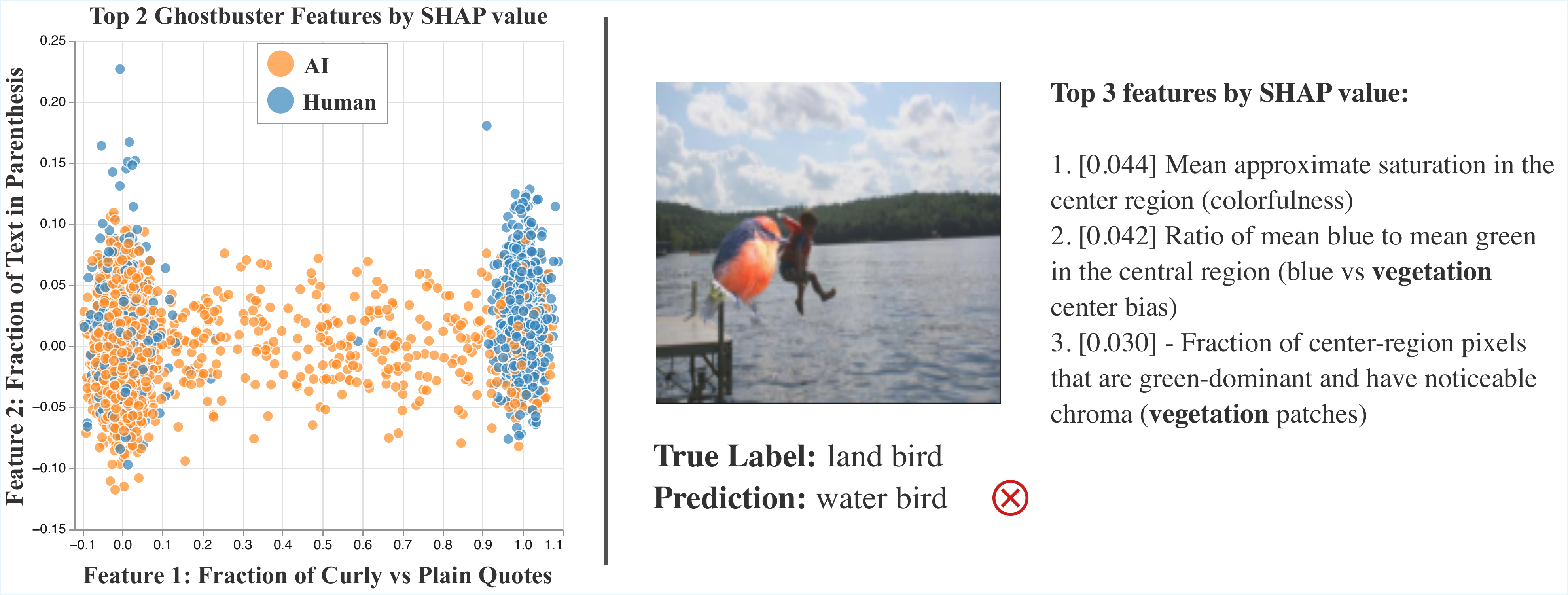}
    \caption{(Left) Distribution of the top 2 LeaPR-learned features with highest SHAP values on the Ghostbuster dataset (gaussian jitter added to aid visualization). On Feature 1 (which measures fraction of ``curly'', or typographic characters, versus plain ASCII quotes), human text tends to cluster at the extremes, while AI-generated text features mid-range values. Feature 2 shows high values primarily for human-written text. (Right) A land bird misclassified by a LeaPR model on the Waterbird dataset; the top SHAP-valued features for this example show a clear reliance on the background.}  
    \label{fig:case-studies}
\end{figure}

\subsection{Case studies: understanding features and model predictions}

The interpretable representations that LeaPR models learn have potential uses beyond their predictive power. We now describe two case studies using SHAP values \citep{lundberg2017unified} as a lens into the patterns that LeaPR models find in their training data, and into why a particular prediction --- especially if erroneous --- was made. SHAP values are a metric of feature importance for a model that can be applied both at a dataset-level or to a particular prediction; we refer to \citet{lundberg2017unified} for details. This is an especially compelling tool for understanding LeaPR models given that their features already come with natural language descriptions, in the form of documentation strings.

First, we compute SHAP values in the Ghostbuster training set to understand what features the LeaPR model trained with D-ID3 and GPT 5-mini has learned to use to identify AI-generated text. Sorting features by their SHAP values on a sample of 150 training examples, we find that two of the top-3 most important features for the model are (1) the ``Fraction of quotation marks that are curly/typographic quotes (e.g., ‘ ’ “ ”) vs plain ASCII quotes, indicating published/edited text'' and the (2) \emph{``Proportion of characters that lie inside parentheses (measures parenthetical/planned content like "(50 words)")''} (the other top-3 feature also looks for kinds of quotation characters and is strongly correlated with the first). Together, these two features already capture distinctive patterns in human- and AI-written samples in Ghostbuster. Figure~\ref{fig:case-studies} (left) shows training samples projected on these two features, with small Gaussian jitter added to both coordinates to aid visualization. For feature 1, human-generated text either has value $1.0$, meaning \emph{all} quote characters are typographical, or ``curly'' quotes, or $0.0$, meaning \emph{all} quotes in the text are plain ASCII quotes (this could, for instance, reflect default settings in the user device, with using curly quotes being much more frequent). This is in stark contrast with AI-generated text, which often mixes both kinds of characters in the same text: AI-generated text displays the full range from $0$ to $1$ in this feature. For Feature 2, humans seem much more likely to wrap a significant fraction of the text in parentheses: almost all samples with value over $0.1$ in this feature (over $10$\% in parenthesis) were human-written. LeaPR models allowed us to quickly discover these patterns without having to formulate specific a priori assumptions: combined, our models contain thousands of automatically generated domain-relevant features, thus serving as a highly useful tool for data understanding.

Finally, we conduct a case study on the Waterbird dataset \citep{sagawa2020waterbirds} showing how programmatic features can serve to debug model failures. This dataset contains images with two classes of birds: land birds and water birds --- which are the two target classification labels. However, a trained classifier might learn instead to rely on the background, rather than use features of the bird itself. The dataset contains a subset of land birds placed on water backgrounds, and vice-versa: typically, classification accuracy drops significantly across groups when models learn to predict bird classes based on the (spuriously correlated) background. When we train a LeaPR model with F2 and GPT 5-mini on Waterbird, it achieves 100\% validation accuracy when evaluated on \emph{land birds on land background}, but it drops to 84\% when evaluated on \emph{land birds on water background}. Again, SHAP values can help us understand why this happens in a particular case. Figure~\ref{fig:case-studies} (right) shows the first validation example of a land bird on water background that is misclassified. When we show the top 3 features by their SHAP values, the second and third features explicitly indicate that their goal is to detect \emph{vegetation} --- a spurious feature. We can indeed find several examples of features that \emph{attempt} to characterize the bird, such as ``Fraction of warm (red-dominant) pixels in the center region (bird color cue)'', that are however ignored by the trained predictor. This example shows how LeaPR models can make their failures transparent. Since model failures often reflect properties of the training data, we believe that LeaPR can serve as debugging tool for both models and datasets, applicable to a wide range of domains.

\vspace{-1em}

\section{Limitations and Conclusion}
\vspace{-1em}
We introduced \emph{Learned Programmatic Representation} models, a class of neural network-free models that combine programmatic LLM-generated features and decision tree predictors. We experiment with chess boards, images and text, seeing encouraging initial results exploring this class of models. Moreover, our learned features tend to be easy to interpret: we explore various examples across each of these domains, including using SHAP values to explain individual predictions.

However, several limitations remain to be tackled by future work. First, our methods do not learn deep hierarchical features: each feature is directly computed from the input, but not from other features. Learning deep feature hierarchies is the main advantage of training \emph{deep} rather than shallow neural networks, and we believe that to be an important capability if LeaPR models are to lead to competitive performance in more complex domains. Moreover, our experiments were still done at a small scale, and there are scalability challenges --- both in representation learning as well as in training predictors --- to be overcome if we want to give competitive predictive performance in data-rich domains, like chess, where neural networks improve predictably with more data and compute. 

Despite these limitations, we find our results encouraging for further exploring novel learning paradigms that yield interpretable models \emph{by construction}, rather than post-hoc. With a rapidly advancing AI toolbox, we can imagine that future tools might allow us to learn interpretable models just as easily as we can train neural networks today, with little to no sacrifice in quality. Overcoming the limitations in the LeaPR paradigm can thus be a path to make this possible.

\subsubsection*{Acknowledgments}

This work has been made possible in part by a gift from the Chan Zuckerberg Initiative Foundation to establish the Kempner Institute for the Study of Natural and Artificial Intelligence at Harvard University. In the earlier stages of the work, GP was also supported by a Stanford Graduate Interdisciplinary Fellowship.

\bibliography{iclr2026_conference}
\bibliographystyle{iclr2026_conference}

\appendix

\section{Implementation details for F2 and D-ID3}
\label{app:practical}

Both F2 and D-ID3 run for a user-specified number of iterations; this number is exactly equal to the number of LLM calls that the algorithm will perform, allowing users to budget for LLM usage. In a sense, both algorithms are ``anytime algorithms'' --- they can always return their latest set of learned features. The algorithms return a \emph{representation}, rather than a predictor (e.g. the decision tree constructed by D-ID3), to allow for separation of concerns: having a representation, users can later iterate on learning predictors (which need not be decision trees) without additional LLM calls. Most of the time in F2 and D-ID3 is generally spent computing features; luckily, feature computation for all relevant examples is embarrassingly parallel, and we exploit this in our implementation. During training, we always validate proposed features on a subset of the training set (we use $10k$ examples in our experiments), and discard features that throw exceptions, timeout, or return non-finite values for some example (e.g., \texttt{NaN} or $\pm{} \infty$).

Our training runs for D-ID3 were the most expensive, with cost ranging from $0.5$ to $5$ US dollars per run with GPT 5-mini (1000 iterations). Runs with F2 were around 10x cheaper, due to performing 10x less LLM calls. Runs took from $5$ to $24$h on a CPU-only commodity machine. 

\section{Experimental Details}

\subsection{Transformer Training}

We followed the architectural hyperparameters in \citet{ruoss2024amortized}, and trained the 270M model with cross-entropy loss on their same tokenization scheme, and the same learning rate of $10^{-5}$. We train for $300k$ steps with a batch size of 400 on a machine with 4x H100 NVIDIA GPUs. This gives around 2.4 epochs over $50M$ data points. We find training to be generally stable, with top-1 move accuracy slowly but monotonically improving across checkpoints (whereas regression metrics, like Pearson correlation, seem less stable, often temporarily decreasing before improving again).

\subsection{MNIST and Fashion-MNIST Training \label{app:mnist}}

We train standard ResNet-50 and EfficientNet-V2 models for 4000 steps and a batch size of 1024 with Adam on a single H100 GPU. We use the default random initialization from PyTorch. For ResNet-50, we use a learning rate of 0.03; for EfficientNet-V2, we use 0.001, which we tuned using the validation set.

\section{Prompts and Example Features}
\label{app:prompts}

\subsection{Chess Position Evaluation}

\subsubsection{F2}
\begin{lstlisting}[language=Python]
You are an expert chess programmer creating feature functions to help a machine learning model predict chess position evaluations.

Your task is to write a feature function that helps discriminate between board positions with different evaluations (e.g., probability that white or black wins).
A feature function is a Python function that takes a chess Board and computes a feature out of the board. It should return a float, but note that a feature could also be effectively boolean-valued (0.0 or 1.0), or integer-valued, even if its type is float.

You have access to the following API from the `chess` library:

# Chess API Documentation
## Class chess.Board Methods
- board.turn: True if White to move, False if Black
- board.fullmove_number: Current move number
- board.halfmove_clock: Halfmove clock for 50-move rule
- board.is_check(): True if current player is in check
- board.is_checkmate(): True if current player is checkmated
- board.is_stalemate(): True if stalemate
- board.is_insufficient_material(): Returns True if insufficient material
- board.piece_at(square): Returns piece at given square (or None)
- board.piece_map(): Returns dict mapping squares to pieces
- board.legal_moves: Iterator over legal moves
- board.attackers(color, square): Returns set of squares attacking given square
- board.is_attacked_by(color, square): Returns True if square is attacked by color

## Chess Squares and Pieces
- chess.A1, chess.A2, ..., chess.H8: Square constants
- chess.PAWN, chess.KNIGHT, chess.BISHOP, chess.ROOK, chess.QUEEN, chess.KING: Piece types
- chess.WHITE, chess.BLACK: Colors
- piece.piece_type: Type of piece (PAWN, KNIGHT, etc.)
- piece.color: Color of piece (WHITE or BLACK)

## Useful Functions
- chess.square_name(square): Convert square index to name (e.g., "e4")
- chess.parse_square(name): Convert square name to index
- chess.square_file(square): Get file (0-7) of square
- chess.square_rank(square): Get rank (0-7) of square
- chess.square_distance(sq1, sq2): Manhattan distance between squares

## Current Feature Database
Here are some of our existing features and their importance to the current model (higher importance means this is a more useful feature for the current model):

Feature:
def feature(board: chess.Board) -> float:
    'King centralization in endgames: positive if White king is closer to center than Black king (only active when low material)'
    total_non_king_pieces = sum(1 for _, p in board.piece_map().items() if p.piece_type != chess.KING)
    # Activation threshold: small material (pawns + pieces <= 6)
    if total_non_king_pieces > 6:
        return 0.0
    center_sq = [chess.parse_square(s) for s in ('d4','e4','d5','e5')]
    center_sq_avg = sum(center_sq)  # not used directly; use center coords (3.5,3.5)
    def king_dist(color):
        for sq, p in board.piece_map().items():
            if p.piece_type == chess.KING and p.color == color:
                # distance to center by min over central squares
                return min(chess.square_distance(sq, c) for c in center_sq)
        return 8.0
    wd = king_dist(chess.WHITE)
    bd = king_dist(chess.BLACK)
    # Normalize in range roughly -1..1
    return float((bd - wd) / 8.0)

Importance: 0.014
---

Feature:
def feature(board: chess.Board) -> float:
    'Bishop-pair and minor-piece balance: (White advantage) bishops and minor piece composition bonus'
    w_bishops = 0
    b_bishops = 0
    w_minors = 0
    b_minors = 0
    for _, p in board.piece_map().items():
        if p.piece_type == chess.BISHOP:
            if p.color == chess.WHITE:
                w_bishops += 1
            else:
                b_bishops += 1
        if p.piece_type in (chess.BISHOP, chess.KNIGHT):
            if p.color == chess.WHITE:
                w_minors += 1
            else:
                b_minors += 1
    score = 0.0
    # bishop pair bonus
    if w_bishops >= 2:
        score += 0.6
    if b_bishops >= 2:
        score -= 0.6
    # minor piece imbalance small weight
    score += 0.12 * (w_minors - b_minors)
    return float(score)

Importance: 0.039
---

Feature:
def feature(board: chess.Board) -> float:
    'Center control: difference in control of d4,e4,d5,e5 (occupied=1, attacked=0.5)'
    center_squares = [chess.parse_square(s) for s in ('d4','e4','d5','e5')]
    def control_for(color):
        c = 0.0
        for sq in center_squares:
            occ = board.piece_at(sq)
            if occ is not None and occ.color == color:
                c += 1.0
            # attacked by color
            if board.is_attacked_by(color, sq):
                c += 0.5
        return c
    wc = control_for(chess.WHITE)
    bc = control_for(chess.BLACK)
    return float(wc - bc)

Importance: 0.044
---

Feature:
def feature(board: chess.Board) -> float:
    'Piece activity squares: difference in number of unique squares attacked by non-pawn, non-king pieces (White - Black) normalized by 64'
    def active_squares(color):
        count = 0
        for sq in range(64):
            attackers = board.attackers(color, sq)
            found = False
            for a in attackers:
                p = board.piece_at(a)
                if p is None:
                    continue
                if p.piece_type not in (chess.PAWN, chess.KING):
                    found = True
                    break
            if found:
                count += 1
        return count
    w = active_squares(chess.WHITE)
    b = active_squares(chess.BLACK)
    return float((w - b) / 64.0)

Importance: 0.329
---

Feature:
def feature(board: chess.Board) -> float:
    'Pawn structure weakness: (Black penalties - White penalties), positive if Black worse (good for White). Penalty = doubled*0.5 + isolated*0.7'
    def pawn_penalty(color):
        files = {f:0 for f in range(8)}
        pawn_sqs = []
        for sq, p in board.piece_map().items():
            if p.piece_type == chess.PAWN and p.color == color:
                f = chess.square_file(sq)
                files[f] += 1
                pawn_sqs.append(sq)
        doubled = sum(max(0, cnt-1) for cnt in files.values())
        isolated = 0
        for sq in pawn_sqs:
            f = chess.square_file(sq)
            if files.get(f-1,0) == 0 and files.get(f+1,0) == 0:
                isolated += 1
        return doubled * 0.5 + isolated * 0.7
    bp = pawn_penalty(chess.BLACK)
    wp = pawn_penalty(chess.WHITE)
    return float(bp - wp)

Importance: 0.065
---

Feature:
def feature(board: chess.Board) -> float:
    'King safety pressure: weighted sum of attackers near each king (positive = pressure on Black king > White king)'
    def king_square(color):
        for sq, p in board.piece_map().items():
            if p.piece_type == chess.KING and p.color == color:
                return sq
        return None
    wk = king_square(chess.WHITE)
    bk = king_square(chess.BLACK)
    values = {chess.PAWN:1.0, chess.KNIGHT:3.0, chess.BISHOP:3.25, chess.ROOK:5.0, chess.QUEEN:9.0, chess.KING:0.5}
    def pressure_on(king_sq, attacker_color):
        if king_sq is None:
            return 0.0
        attackers = board.attackers(attacker_color, king_sq)
        s = 0.0
        for a in attackers:
            p = board.piece_at(a)
            if p is None:
                continue
            dist = chess.square_distance(a, king_sq)
            s += values.get(p.piece_type, 0.0) / (1.0 + dist)
        return s
    p_on_black = pressure_on(bk, chess.WHITE)
    p_on_white = pressure_on(wk, chess.BLACK)
    return float(p_on_black - p_on_white)

Importance: 0.005
---

Feature:
def feature(board: chess.Board) -> float:
    'Undefended high-value threats: (value of Black pieces under more attackers than defenders) - (value of White pieces similarly threatened)'
    values = {chess.PAWN:1.0, chess.KNIGHT:3.0, chess.BISHOP:3.25, chess.ROOK:5.0, chess.QUEEN:9.0, chess.KING:0.0}
    threat_black = 0.0
    threat_white = 0.0
    for sq, p in board.piece_map().items():
        if p.piece_type == chess.PAWN:
            continue
        attackers = board.attackers(not p.color, sq)
        defenders = board.attackers(p.color, sq)
        atk = sum(1 for _ in attackers)
        defn = sum(1 for _ in defenders)
        if atk > defn and atk > 0:
            score = values.get(p.piece_type, 0.0) * (atk - defn)
            if p.color == chess.BLACK:
                threat_black += score
            else:
                threat_white += score
    return float(threat_black - threat_white)

Importance: 0.045
---

Feature:
def feature(board: chess.Board) -> float:
    'Material balance (White minus Black) using common piece values: P=1,N=3,B=3.25,R=5,Q=9'
    values = {chess.PAWN:1.0, chess.KNIGHT:3.0, chess.BISHOP:3.25, chess.ROOK:5.0, chess.QUEEN:9.0, chess.KING:0.0}
    total = 0.0
    for sq, piece in board.piece_map().items():
        val = values.get(piece.piece_type, 0.0)
        total += val if piece.color == chess.WHITE else -val
    return float(total)

Importance: 0.199
---

Feature:
def feature(board: chess.Board) -> float:
    'Normalized mobility difference: (White legal moves - Black legal moves) / 100'
    try:
        white_moves = 0
        black_moves = 0
        # count current side moves
        white_board = board.copy()
        white_board.turn = chess.WHITE
        white_moves = sum(1 for _ in white_board.legal_moves)
        black_board = board.copy()
        black_board.turn = chess.BLACK
        black_moves = sum(1 for _ in black_board.legal_moves)
        return float((white_moves - black_moves) / 100.0)
    except Exception:
        return 0.0

Importance: 0.165
---

Feature:
def feature(board: chess.Board) -> float:
    'Passed pawns score: sum of passed-pawn strengths (White minus Black), advanced pawns weighted more'
    def is_passed(sq, color):
        f = chess.square_file(sq)
        r = chess.square_rank(sq)
        if color == chess.WHITE:
            ahead_ranks = range(r+1, 8)
            opp_color = chess.BLACK
            for ar in ahead_ranks:
                for df in (-1,0,1):
                    ff = f + df
                    if 0 <= ff < 8:
                        sq2 = chess.square(ff, ar)
                        p = board.piece_at(sq2)
                        if p is not None and p.color == opp_color and p.piece_type == chess.PAWN:
                            return False
            return True
        else:
            ahead_ranks = range(r-1, -1, -1)
            opp_color = chess.WHITE
            for ar in ahead_ranks:
                for df in (-1,0,1):
                    ff = f + df
                    if 0 <= ff < 8:
                        sq2 = chess.square(ff, ar)
                        p = board.piece_at(sq2)
                        if p is not None and p.color == opp_color and p.piece_type == chess.PAWN:
                            return False
            return True
    score_w = 0.0
    score_b = 0.0
    for sq, p in board.piece_map().items():
        if p.piece_type != chess.PAWN:
            continue
        rank = chess.square_rank(sq)
        if p.color == chess.WHITE:
            if is_passed(sq, chess.WHITE):
                advancement = (rank - 1) / 6.0 if rank >= 1 else 0.0
                score_w += 1.0 + advancement
        else:
            if is_passed(sq, chess.BLACK):
                advancement = (6 - rank) / 6.0 if rank <= 6 else 0.0
                score_b += 1.0 + advancement
    return float(score_w - score_b)

Importance: 0.094
---

# Task
Generate 10 new chess board feature functions in Python that:

1. Help us discriminate between strong and weak board positions, hopefully with positions before and after the optimal split point having the lowest possible variance between their evaluations.
2. Return a float value given a board position.
3. Handle edge cases gracefully - won't crash on unusual positions

Your task is to generate diverse, creative features that are relevant to explain the evaluations for the board positions shown above. Focus on features that would help distinguish between positions of different strengths. These features will be used in this decision tree that will predict the evaluation of a given board position in estimated % win probability for white (e.g., 20 means Black winning with around 80% probability). Think about new features that would help such a predictor in the particular cases above, trying to add information that the already existing features shown above are missing.

# Code Requirements

- Use single quotes for docstrings: "description here"
- No markdown code blocks
- No explanatory text after the function
- Each function should be complete and standalone, and return a float

# Output Format
Generate exactly 10 features in this format:

def feature(board: chess.Board) -> float:
    "Simple, clear description of what this feature measures"
    # ... Calculate and return the feature value
    return result

def feature(board: chess.Board) -> float:
    "Another feature description"
    # ... Calculate and return the feature value
    return result

The body of the function can be anything, but the first line (function declaration) should be identical to those examples above, and the second line should be a one-line docstring. Don't output explanatory text - just the function definitions as shown above.

Optimize for producing discriminant features that are novel compared to the existing features and that are likely to achieve a high importance for scoring positions, once we retrain the model using your new features combined with the existing ones.

\end{lstlisting}

\subsubsection{D-ID3 - Prompt}
\begin{lstlisting}[language=Python]
You are an expert chess programmer creating feature functions to help a machine learning model predict chess position evaluations.

Your task is to write a feature function that helps discriminate between the board positions given below.
A feature function is a Python function that takes a chess Board and computes a feature out of the board. It should return a float, but note that a feature could also be effectively boolean-valued (0.0 or 1.0), or integer-valued, even if its type is float.

You have access to the following API from the `chess` library:

# Chess API Documentation
## Class chess.Board Methods
- board.turn: True if White to move, False if Black
- board.fullmove_number: Current move number
- board.halfmove_clock: Halfmove clock for 50-move rule
- board.is_check(): True if current player is in check
- board.is_checkmate(): True if current player is checkmated
- board.is_stalemate(): True if stalemate
- board.is_insufficient_material(): Returns True if insufficient material
- board.piece_at(square): Returns piece at given square (or None)
- board.piece_map(): Returns dict mapping squares to pieces
- board.legal_moves: Iterator over legal moves
- board.attackers(color, square): Returns set of squares attacking given square
- board.is_attacked_by(color, square): Returns True if square is attacked by color

## Chess Squares and Pieces
- chess.A1, chess.A2, ..., chess.H8: Square constants
- chess.PAWN, chess.KNIGHT, chess.BISHOP, chess.ROOK, chess.QUEEN, chess.KING: Piece types
- chess.WHITE, chess.BLACK: Colors
- piece.piece_type: Type of piece (PAWN, KNIGHT, etc.)
- piece.color: Color of piece (WHITE or BLACK)

## Useful Functions
- chess.square_name(square): Convert square index to name (e.g., "e4")
- chess.parse_square(name): Convert square name to index
- chess.square_file(square): Get file (0-7) of square
- chess.square_rank(square): Get rank (0-7) of square
- chess.square_distance(sq1, sq2): Manhattan distance between squares

# Task
Generate 10 new chess board feature functions in Python that:

1. Help us discriminate between strong and weak board positions, hopefully with positions before and after the optimal split point having the lowest possible variance between their evaluations.
2. Return a float value given a board position.
3. Handle edge cases gracefully - won't crash on unusual positions

Your task is to generate diverse, creative features that are relevant to explain the evaluations for the board positions shown above. Focus on features that would help distinguish between positions of different strengths. These features will be used in this decision tree that will predict the evaluation of a given board position in estimated % win probability for white (e.g., 20 means Black winning with around 80% probability). Think about new features that would help such a predictor in the particular cases above, trying to add information that the already existing features shown above are missing.

# Code Requirements

- Use single quotes for docstrings: "description here"
- No markdown code blocks
- No explanatory text after the function
- Each function should be complete and standalone, and return a float

# Output Format
Generate exactly 10 features in this format:

def feature(board: chess.Board) -> float:
    "Simple, clear description of what this feature measures"
    # ... Calculate and return the feature value
    return result

def feature(board: chess.Board) -> float:
    "Another feature description"
    # ... Calculate and return the feature value
    return result

The body of the function can be anything, but the first line (function declaration) should be identical to those examples above, and the second line should be a one-line docstring. Don't output explanatory text - just the function definitions as shown above.

# Current decision tree node
You are currently focusing on features that explain the position evaluation of board positions in the following subtree of a decision tree:

[root]
 -> value < 3.000 for "Measure the material balance between both players"  -> value < 1.182 for "Counts the number of pieces for each player and computes the ratio of the piece counts."  -> value > 0.650 for "Counts the number of squares attacked by pieces of each color to assess control of the board." 

# Board positions
Here are examples of board positions in this subtree, along with their position evaluations (computed by Stockfish):

Board:
. . r r . . k .
. . . . . p p p
p b . . p . . .
. p . . P . . .
. P . . N P n q
P . . . . . P .
. B . . Q . . P
R . . . . R . K
Evaluation: 38.08281678856247
---

Board:
r . . . . . k .
. b . r . p p .
p . . P . . . .
. p . . . . . Q
. P . b . . . .
P . . . . . . .
. . . . . . P P
. . . . . R . K
Evaluation: 15.383900101915987
---

Board:
r n b . k . . r
p p p . n p p p
. . . b . . . .
. . . B . . . .
. . . P P p . q
. . . . . . . .
P P P . . . P P
R N B Q . K N R
Evaluation: 47.700323643230064
---

Board:
. . . r . . . k
p p . . . p p .
. . . . p . . p
. . b r P . . P
. q . N R . . .
. . . . B . . .
. . . . . K P .
. . R Q . . . .
Evaluation: 50.0
---

Board:
r n b q k b n r
p p p p . p p p
. . . . . . . .
. . . . p . . .
. . . . P . . .
. . . . . N . .
P P P P . P P P
R N B Q K B . R
Evaluation: 51.19644756166626
---

Optimize for producing discriminant features that are novel compared to the existing features used to arrive at this subtree. Focus on explaining the differences between the board positions shown above.
\end{lstlisting}

\subsubsection{D-ID3 - Example Features}
\begin{lstlisting}[language=Python]
def feature(board: chess.Board) -> float:
    "Measure the material balance between both players"
    material_values = {chess.PAWN: 1, chess.KNIGHT: 3, chess.BISHOP: 3, chess.ROOK: 5, chess.QUEEN: 9, chess.KING: 0}
    white_material = sum(material_values[piece.piece_type] for piece in board.piece_map().values() if piece.color == chess.WHITE)
    black_material = sum(material_values[piece.piece_type] for piece in board.piece_map().values() if piece.color == chess.BLACK)
    return float(white_material - black_material)

def feature(board: chess.Board) -> float:
    "Counts the number of pieces for each player and computes the ratio of the piece counts."
    piece_count_white = sum(1 for piece in board.piece_map().values() if piece.color == chess.WHITE)
    piece_count_black = sum(1 for piece in board.piece_map().values() if piece.color == chess.BLACK)
    if piece_count_black == 0:
        return float('inf')  # Black has no pieces left
    return piece_count_white / piece_count_black

def feature(board: chess.Board) -> float:
    "Counts the number of squares attacked by pieces of each color to assess control of the board."
    white_attacks = sum(board.is_attacked_by(chess.WHITE, square) for square in chess.SQUARES)
    black_attacks = sum(board.is_attacked_by(chess.BLACK, square) for square in chess.SQUARES)
    control_ratio = white_attacks / (black_attacks + 1)  # Avoid division by zero
    return float(control_ratio)

def feature(board: chess.Board) -> float:
    "Calculates the center control by counting pieces in the central squares."
    central_squares = [chess.D4, chess.D5, chess.E4, chess.E5]
    control = sum(1 for square in central_squares if board.piece_at(square) is not None)
    return float(control)

def feature(board: chess.Board) -> float:
    "Calculates the total piece value for each player based on standard chess piece values."
    piece_values = {
        chess.PAWN: 1,
        chess.KNIGHT: 3,
        chess.BISHOP: 3,
        chess.ROOK: 5,
        chess.QUEEN: 9,
        chess.KING: 0  # King is invaluable
    }
    white_value = sum(piece_values[piece.piece_type] for piece in board.piece_map().values() if piece.color == chess.WHITE)
    black_value = sum(piece_values[piece.piece_type] for piece in board.piece_map().values() if piece.color == chess.BLACK)
    return float(white_value - black_value)

\end{lstlisting}

\subsection{Image Classification}

\subsubsection{F2 }
\begin{lstlisting}[language=Python]
You are an expert computer vision programmer creating evaluation features for a machine learning model that predicts values from images.

This is a classification task with the following classes: 0: landbird, 1: waterbird.

Your task is to write a feature function that helps discriminate between the image classes above.
A feature function is a Python function that takes an image array and computes a feature out of the image. It should return a float, but note that a feature could also be effectively boolean-valued (0.0 or 1.0), or integer-valued, even if its type is float.

You have access to the following API from image processing libraries:

# Image Processing API Documentation

The features receive an image as a numpy array, so you can use any numpy functions on it. For RGB images, shape is (height, width, 3). For grayscale, shape is (height, width).

## Image Processing Methods
- image.shape: Returns (height, width, channels) for RGB or (height, width) for grayscale
- image.mean(): Average pixel intensity across all channels
- image.std(): Standard deviation of pixel intensities
- image.max(), image.min(): Maximum and minimum pixel values
- np.sum(image): Sum of all pixel values
- np.count_nonzero(image): Count of non-zero pixels

## Handle Both Grayscale and RGB
- Check format: len(image.shape) == 2 for grayscale, len(image.shape) == 3 for RGB
- Unpack safely: h, w = image.shape[:2]  # Works for both formats
- For RGB only: image[:,:,0] (red), image[:,:,1] (green), image[:,:,2] (blue)

## Useful Functions
- np.mean(image): Average intensity
- np.std(image): Standard deviation
- np.gradient(image): Image gradients - for RGB use on single channel: np.gradient(image[:,:,0])
- np.where(condition, x, y): Conditional selection
- np.argmax(image), np.argmin(image): Location of max/min values
- np.percentile(image, q): Percentile values
- np.histogram(image.flatten(), bins): Intensity histogram

## Spatial Analysis
- image[start_row:end_row, start_col:end_col]: Region selection
- Center region: image[h//4:3*h//4, w//4:3*w//4]
- Edge detection: np.gradient(np.mean(image, axis=2)) for RGB
- Color channel differences: image[:,:,0] - image[:,:,1]

## Example Feature Function
def feature(image: np.ndarray) -> float:
    "Average pixel intensity in the center region"
    if len(image.shape) == 3:
        h, w, c = image.shape
        gray = np.mean(image, axis=2)
    else:
        h, w = image.shape
        gray = image
    center_h, center_w = h // 4, w // 4
    center_region = gray[center_h:3*center_h, center_w:3*center_w]
    return float(np.mean(center_region))

## Current Feature Database
Here are some existing features and their importances to the current classifier (importance = benefit from that feature, higher is better):

Feature:
def feature(image: np.ndarray) -> float:
    'Relative difference in average blue intensity between bottom half and top half'
    import numpy as np
    h, w = image.shape[:2]
    if len(image.shape) != 3 or image.shape[2] < 3:
        return float(0.0)
    b = image[:, :, 2].astype(float)
    top_mean = np.mean(b[:h // 2, :]) if h // 2 > 0 else np.mean(b)
    bot_mean = np.mean(b[h // 2:, :]) if h - h // 2 > 0 else np.mean(b)
    denom = (np.mean(b) + 1e-8)
    return float((bot_mean - top_mean) / denom)

Importance: 0.081
---

Feature:
def feature(image: np.ndarray) -> float:
    'Aspect ratio (width/height) of bounding box of pixels significantly different from median intensity'
    import numpy as np
    h, w = image.shape[:2]
    if len(image.shape) == 3:
        gray = np.mean(image, axis=2).astype(float)
    else:
        gray = image.astype(float)
    med = np.median(gray)
    dynamic_range = np.max(gray) - np.min(gray)
    thresh = 0.15 * (dynamic_range + 1e-8)
    mask = np.abs(gray - med) > thresh
    if not np.any(mask):
        return float(0.5)
    rows = np.where(np.any(mask, axis=1))[0]
    cols = np.where(np.any(mask, axis=0))[0]
    if rows.size == 0 or cols.size == 0:
        return float(0.5)
    r0, r1 = rows[0], rows[-1]
    c0, c1 = cols[0], cols[-1]
    bbox_h = (r1 - r0 + 1)
    bbox_w = (c1 - c0 + 1)
    return float(bbox_w / (bbox_h + 1e-8))

Importance: 0.082
---

Feature:
def feature(image: np.ndarray) -> float:
    'Left-right symmetry score (1.0 = perfectly symmetric, lower = less symmetric)'
    import numpy as np
    if len(image.shape) == 3:
        gray = np.mean(image, axis=2).astype(float)
    else:
        gray = image.astype(float)
    flipped = np.fliplr(gray)
    diff = np.mean(np.abs(gray - flipped))
    # normalize by image contrast to avoid small-image issues
    denom = np.mean(np.abs(gray - np.mean(gray))) + 1e-8
    norm_diff = diff / denom
    score = 1.0 - np.tanh(norm_diff)  # bounded between ~0 and 1
    return float(np.clip(score, 0.0, 1.0))

Importance: 0.084
---

Feature:
def feature(image: np.ndarray) -> float:
    'Ratio of average horizontal gradient magnitude to vertical gradient magnitude (texture orientation)'
    import numpy as np
    # compute gray
    if len(image.shape) == 3:
        gray = np.mean(image, axis=2).astype(float)
    else:
        gray = image.astype(float)
    gy, gx = np.gradient(gray)
    mean_dx = np.mean(np.abs(gx))
    mean_dy = np.mean(np.abs(gy))
    return float(mean_dx / (mean_dy + 1e-8))

Importance: 0.159
---

Feature:
def feature(image: np.ndarray) -> float:
    'Proportion of pixels where green channel is significantly high (vegetation cue)'
    import numpy as np
    h, w = image.shape[:2]
    if len(image.shape) != 3 or image.shape[2] < 3:
        return float(0.0)
    img = image.astype(float)
    r, g, b = img[:, :, 0], img[:, :, 1], img[:, :, 2]
    # require green greater than both red and blue and at least above 60th percentile
    thresh = np.percentile(g.flatten(), 60)
    mask = (g > r) & (g > b) & (g > thresh)
    return float(np.count_nonzero(mask) / (h * w + 1e-12))

Importance: 0.094
---

Feature:
def feature(image: np.ndarray) -> float:
    'Fraction of pixels brighter than the 90th percentile (bright spot proportion)'
    import numpy as np
    if len(image.shape) == 3:
        gray = np.mean(image, axis=2).astype(float)
    else:
        gray = image.astype(float)
    thresh = np.percentile(gray.flatten(), 90)
    frac = np.count_nonzero(gray > thresh) / (gray.size + 1e-12)
    return float(frac)

Importance: 0.080
---

Feature:
def feature(image: np.ndarray) -> float:
    'Normalized difference between center region brightness and border brightness'
    import numpy as np
    h, w = image.shape[:2]
    if len(image.shape) == 3:
        gray = np.mean(image, axis=2).astype(float)
    else:
        gray = image.astype(float)
    ch0, ch1 = h // 4, w // 4
    center = gray[ch0:3 * ch0 or h, ch1:3 * ch1 or w]
    # border defined as whole minus center
    mask_border = np.ones_like(gray, dtype=bool)
    mask_border[ch0:3 * ch0 or h, ch1:3 * ch1 or w] = False
    center_mean = np.mean(center) if center.size > 0 else 0.0
    border_mean = np.mean(gray[mask_border]) if np.any(mask_border) else 0.0
    denom = np.mean(np.abs(gray)) + 1e-8
    return float((center_mean - border_mean) / denom)

Importance: 0.088
---

Feature:
def feature(image: np.ndarray) -> float:
    'Proportion of pixels where blue channel is dominant (blue > red and blue > green)'
    import numpy as np
    h, w = image.shape[:2]
    if len(image.shape) != 3 or image.shape[2] < 3:
        return float(0.0)
    img = image.astype(float)
    r, g, b = img[:, :, 0], img[:, :, 1], img[:, :, 2]
    mask = (b > r) & (b > g)
    return float(np.count_nonzero(mask) / (h * w + 1e-12))

Importance: 0.117
---

Feature:
def feature(image: np.ndarray) -> float:
    'Edge density: fraction of pixels with gradient magnitude above (mean+std)'
    import numpy as np
    if len(image.shape) == 3:
        gray = np.mean(image, axis=2).astype(float)
    else:
        gray = image.astype(float)
    gy, gx = np.gradient(gray)
    mag = np.hypot(gx, gy)
    thresh = np.mean(mag) + np.std(mag)
    count = np.count_nonzero(mag > thresh)
    total = gray.size
    return float(count / (total + 1e-12))

Importance: 0.114
---

Feature:
def feature(image: np.ndarray) -> float:
    'Average per-pixel color "saturation" approximated by (max-min)/max across channels'
    import numpy as np
    if len(image.shape) != 3 or image.shape[2] < 3:
        return float(0.0)
    img = image.astype(float)
    mx = np.max(img, axis=2)
    mn = np.min(img, axis=2)
    # avoid division by zero
    sat = (mx - mn) / (mx + 1e-8)
    return float(np.mean(sat))

Importance: 0.100
---

## Task
Generate 10 NEW image features that:

1. Are different from existing features
2. Capture useful visual patterns
3. Return float values
4. Handle edge cases gracefully - Won't crash on unusual images
5. Use simple, short docstrings - Use single quotes, not triple quotes
6. Are efficient to compute

Your task is to generate diverse, creative features that capture different aspects of image content for prediction. Focus on features that would help distinguish between different samples. These features will be used as input to a learned model that predicts target values from images.

## IMPORTANT CODE REQUIREMENTS
- Use SINGLE quotes for docstrings: "description here"
- NO triple quotes (""") anywhere in the code
- NO markdown code blocks
- NO explanatory text after the function
- Each function should be complete and standalone

## Output Format
Generate exactly 10 features in this format:

def feature(image: np.ndarray) -> float:
    "Clear description of what this feature measures"
    # ... Calculate and return the feature value
    return float(result)

def feature(image: np.ndarray) -> float:
    "Another feature description"
    # ... Calculate and return the feature value
    return float(result)

The body of the functions can be anything, but the first line (function declaration) should be identical to those examples above (always 'def feature(...)'), and the second line should be a one-line docstring. Don't output explanatory text - just the function definitions as shown above.
\end{lstlisting}

\subsubsection{D-ID3 - Prompt}
\begin{lstlisting}[language=Python]
You are an expert image processing programmer creating feature functions to help a machine learning model perform image classification.

This is a classification task with the following classes: 0: digit zero, 1: digit one, 2: digit two, 3: digit three, 4: digit four, 5: digit five, 6: digit six, 7: digit seven, 8: digit eight, 9: digit nine.

Your task is to write a feature function that helps discriminate between the image samples given below.
A feature function is a Python function that takes an image array and computes a feature out of the image. It should return a float, but note that a feature could also be effectively boolean-valued (0.0 or 1.0), or integer-valued, even if its type is float.

You have access to the following API from image processing libraries:

# Image Processing API Documentation

The features receive an image as a numpy array, so you can use any numpy functions on it. For RGB images, shape is (height, width, 3). For grayscale, shape is (height, width).

## Image Processing Methods
- image.shape: Returns (height, width, channels) for RGB or (height, width) for grayscale
- image.mean(): Average pixel intensity across all channels
- image.std(): Standard deviation of pixel intensities
- image.max(), image.min(): Maximum and minimum pixel values
- np.sum(image): Sum of all pixel values
- np.count_nonzero(image): Count of non-zero pixels

## Handle Both Grayscale and RGB
- Check format: len(image.shape) == 2 for grayscale, len(image.shape) == 3 for RGB
- Unpack safely: h, w = image.shape[:2]  # Works for both formats
- For RGB only: image[:,:,0] (red), image[:,:,1] (green), image[:,:,2] (blue)

## Useful Functions
- np.mean(image): Average intensity
- np.std(image): Standard deviation
- np.gradient(image): Image gradients - for RGB use on single channel: np.gradient(image[:,:,0])
- np.where(condition, x, y): Conditional selection
- np.argmax(image), np.argmin(image): Location of max/min values
- np.percentile(image, q): Percentile values
- np.histogram(image.flatten(), bins): Intensity histogram

## Spatial Analysis
- image[start_row:end_row, start_col:end_col]: Region selection
- Center region: image[h//4:3*h//4, w//4:3*w//4]
- Edge detection: np.gradient(np.mean(image, axis=2)) for RGB
- Color channel differences: image[:,:,0] - image[:,:,1]

## Example Feature Function
def feature(image: np.ndarray) -> float:
    "Average pixel intensity in the center region"
    if len(image.shape) == 3:
        h, w, c = image.shape
        gray = np.mean(image, axis=2)
    else:
        h, w = image.shape
        gray = image
    center_h, center_w = h // 4, w // 4
    center_region = gray[center_h:3*center_h, center_w:3*center_w]
    return float(np.mean(center_region))

# Task
Generate 10 new image feature functions in Python that:

1. Help us discriminate between different image classes, hopefully with samples before and after the optimal split point having the lowest possible variance between their classifications.
2. Return a float value given an image.
3. Handle edge cases gracefully - won't crash on unusual images

Your task is to generate diverse, creative features that are relevant to explain the classifications for the image samples shown above. Focus on features that would help distinguish between samples of different classes. These features will be used in this decision tree that will predict the classification of a given image sample. Think about new features that would help such a predictor in the particular cases above, trying to add information that the already existing features shown above are missing.

# Code Requirements
- Use single quotes for docstrings: "description here"
- No markdown code blocks
- No explanatory text after the function
- Each function should be complete and standalone, and return a float

# Output Format
Generate exactly 10 features in this format:

def feature(image: np.ndarray) -> float:
    "Simple, clear description of what this feature measures"
    # ... Calculate and return the feature value
    return result

def feature(image: np.ndarray) -> float:
    "Another feature description"
    # ... Calculate and return the feature value
    return result

The body of the function can be anything, but the first line (function declaration) should be identical to those examples above, and the second line should be a one-line docstring. Don't output explanatory text - just the function definitions as shown above.

# Current decision tree node
You are currently focusing on features that explain the image classifications in the following subtree of a decision tree:

[root]
 -> value > 0.344 for "Ratio of edge pixels to the total number of pixels"  -> value > 72.865 for "Computes the brightness of the central region as a percentage of the whole image"  -> value > 0.805 for "Determines the ratio of vertical to horizontal gradients in the image, indicating edge direction" 

# Image samples
Here are examples of image samples in this subtree, along with their target classifications:

Sample (Target: 9)
Sample (Target: 6)
Sample (Target: 7)
Sample (Target: 4)
Sample (Target: 4)

Optimize for producing discriminant features that are novel compared to the existing features used to arrive at this subtree. Focus on explaining the differences between the image samples shown above.
\end{lstlisting}

\subsubsection{D-ID3 - Example Features}
\begin{lstlisting}[language=Python]
def feature(image: np.ndarray) -> float:
    "Ratio of edge pixels to the total number of pixels"
    if len(image.shape) == 3:
        gray = np.mean(image, axis=2)
    else:
        gray = image
    edges = np.gradient(gray.astype(float))
    edge_pixels = np.count_nonzero(edges[0]) + np.count_nonzero(edges[1])
    total_pixels = gray.size
    return float(edge_pixels / total_pixels)

def feature(image: np.ndarray) -> float:
    "Computes the brightness of the central region as a percentage of the whole image"
    if len(image.shape) == 3:
        gray = np.mean(image, axis=2)
    else:
        gray = image
    h, w = gray.shape
    center_region = gray[h//4:3*h//4, w//4:3*w//4]
    total_brightness = np.sum(gray)
    center_brightness = np.sum(center_region)
    return float(center_brightness) / (total_brightness + 1e-7) * 100

def feature(image: np.ndarray) -> float:
    "Determines the ratio of vertical to horizontal gradients in the image, indicating edge direction"
    if len(image.shape) == 3:
        gray = np.mean(image, axis=2)
    else:
        gray = image
    grad_y, grad_x = np.gradient(gray)
    vertical_grad = np.sum(np.abs(grad_x))
    horizontal_grad = np.sum(np.abs(grad_y))
    return float(vertical_grad) / (horizontal_grad + 1e-7)  # Avoid division by zero

def feature(image: np.ndarray) -> float:
    "Calculates the ratio of bright pixels (above a threshold) to total pixels in the image"
    if len(image.shape) == 3:
        gray = np.mean(image, axis=2)
    else:
        gray = image
    threshold = 200  # A threshold for bright pixels
    bright_pixels = np.count_nonzero(gray > threshold)
    total_pixels = gray.size
    return float(bright_pixels) / total_pixels

def feature(image: np.ndarray) -> float:
    "Measures the contrast of the image based on the standard deviation of pixel intensities"
    if len(image.shape) == 3:
        gray = np.mean(image, axis=2)
    else:
        gray = image
    return float(np.std(gray))
\end{lstlisting}

\subsection{Text Classification}

\subsubsection{F2 }
\begin{lstlisting}[language=Python]
You are an expert text analysis programmer creating evaluation features for a machine learning model that classifies text.

# Text Processing API Documentation

The features receive text as a string, so you can use any string methods and text processing functions.

## String Methods
- text.lower(), text.upper(): Case conversion
- text.strip(): Remove whitespace
- text.split(delimiter): Split into list
- text.count(substring): Count occurrences
- text.startswith(prefix), text.endswith(suffix): Check prefixes/suffixes
- text.find(substring): Find position of substring
- text.replace(old, new): Replace text

## Text Analysis
- len(text): Length of text
- text.isdigit(), text.isalpha(), text.isalnum(): Character type checks
- sum(1 for c in text if c.isupper()): Count uppercase letters
- text.split(): Split on whitespace to get words

## Regular Expressions (re module)
- re.findall(pattern, text): Find all matches
- re.search(pattern, text): Find first match
- re.sub(pattern, replacement, text): Replace patterns
- len(re.findall(r'\w+', text)): Count words
- len(re.findall(r'[.!?]', text)): Count sentences

## Useful Patterns
- Word count: len(text.split())
- Sentence count: text.count('.') + text.count('!') + text.count('?')
- Average word length: sum(len(word) for word in text.split()) / len(text.split())
- Punctuation density: sum(1 for c in text if not c.isalnum() and not c.isspace()) / len(text)

## Example Feature Function
def feature(text: str) -> float:
    "Average word length in the text"
    words = text.split()
    if not words:
        return 0.0
    return sum(len(word) for word in words) / len(words)

## Current Feature Database
Here are some existing features and their performance (Performance improvement = benefit from that feature, higher is better):

Feature:
def feature(text: str) -> float:
    'Average number of words per sentence (sentences split on . ! ?)'
    import re
    if not text or not text.strip():
        return float(0.0)
    # Split on one or more sentence-ending punctuation and filter empties
    sentences = [s.strip() for s in re.split(r'[.!?]+', text) if s.strip()]
    if not sentences:
        return float(0.0)
    total_words = sum(len(s.split()) for s in sentences)
    return float(total_words / len(sentences))

Importance: 0.057
---

Feature:
def feature(text: str) -> float:
    'Average character-uniqueness per word (unique chars / word length), averaged over words'
    words = [w for w in text.split() if any(ch.isalnum() for ch in w)]
    if not words:
        return float(0.0)
    ratios = []
    for w in words:
        chars = [c for c in w if not c.isspace()]
        if not chars:
            continue
        unique = len(set(chars))
        ratios.append(unique / len(chars))
    if not ratios:
        return float(0.0)
    return float(sum(ratios) / len(ratios))

Importance: 0.057
---

Feature:
def feature(text: str) -> float:
    'Proportion of alphabetic characters that are uppercase'
    if not text:
        return float(0.0)
    alpha_chars = [c for c in text if c.isalpha()]
    if not alpha_chars:
        return float(0.0)
    upper_count = sum(1 for c in alpha_chars if c.isupper())
    return float(upper_count / len(alpha_chars))

Importance: 0.083
---

Feature:
def feature(text: str) -> float:
    'Proportion of long words (length > 7) among all words'
    words = [w for w in text.split() if w]
    if not words:
        return float(0.0)
    long_count = sum(1 for w in words if len(w) > 7)
    return float(long_count / len(words))

Importance: 0.194
---

Feature:
def feature(text: str) -> float:
    'Punctuation characters per word (punctuation = not alnum and not whitespace)'
    if not text or not text.strip():
        return float(0.0)
    words = text.split()
    if not words:
        return float(0.0)
    punct_count = sum(1 for c in text if not c.isalnum() and not c.isspace())
    return float(punct_count / len(words))

Importance: 0.096
---

Feature:
def feature(text: str) -> float:
    'Proportion of sentences that are questions (based on ? count over total sentence terminators)'
    if not text or not text.strip():
        return float(0.0)
    question_marks = text.count('?')
    sentence_terminators = text.count('.') + text.count('!') + text.count('?')
    if sentence_terminators == 0:
        return float(0.0)
    return float(question_marks / sentence_terminators)

Importance: 0.023
---

Feature:
def feature(text: str) -> float:
    'Type-token ratio: unique word tokens / total words (case-insensitive, alphanumeric tokens)'
    import re
    tokens = re.findall(r'\w+', text.lower())
    if not tokens:
        return float(0.0)
    unique = len(set(tokens))
    return float(unique / len(tokens))

Importance: 0.089
---

Feature:
def feature(text: str) -> float:
    'Ratio of tokens that contain at least one digit'
    tokens = text.split()
    if not tokens:
        return float(0.0)
    num_with_digit = sum(1 for t in tokens if any(ch.isdigit() for ch in t))
    return float(num_with_digit / len(tokens))

Importance: 0.165
---

Feature:
def feature(text: str) -> float:
    'Longest run of the same character normalized by text length'
    if not text:
        return float(0.0)
    max_run = 1
    current_run = 1
    prev = text[0]
    for c in text[1:]:
        if c == prev:
            current_run += 1
            if current_run > max_run:
                max_run = current_run
        else:
            current_run = 1
            prev = c
    return float(max_run / max(1, len(text)))

Importance: 0.126
---

Feature:
def feature(text: str) -> float:
    'Stopword density: fraction of tokens that are common English stopwords'
    stopwords = {
        'the','and','is','in','it','of','to','a','an','that','this','for','on','with',
        'as','by','at','from','or','be','are','was','were','has','have','not','but',
        'they','their','you','I'
    }
    tokens = [t.lower().strip(".,!?;:\"'()[]") for t in text.split()]
    if not tokens:
        return float(0.0)
    stop_count = sum(1 for t in tokens if t and t in stopwords)
    return float(stop_count / len(tokens))

Importance: 0.111
---

## Task
Generate 10 NEW text features that:

1. Are different from existing features
2. Capture useful textual patterns
3. Return float values
4. Handle edge cases gracefully - Won't crash on unusual texts
5. Use simple, short docstrings - Use single quotes, not triple quotes
6. Are efficient to compute

Your task is to generate diverse, creative features that capture different aspects of text content for classification. Focus on features that would help distinguish between different text classes. These features will be used as input to a learned model that predicts target values from text.

## IMPORTANT CODE REQUIREMENTS
- Use SINGLE quotes for docstrings: "description here"
- NO triple quotes (""") anywhere in the code
- NO markdown code blocks
- NO explanatory text after the function
- Each function should be complete and standalone

## Output Format
Generate exactly 10 features in this format:

def feature(text: str) -> float:
    "Clear description of what this feature measures"
    # ... Calculate and return the feature value
    return float(result)

def feature(text: str) -> float:
    "Another feature description"
    # ... Calculate and return the feature value
    return float(result)

The body of the functions can be anything, but the first line (function declaration) should be identical to those examples above (always 'def feature(...)'), and the second line should be a one-line docstring. Don't output explanatory text - just the function definitions as shown above.

\end{lstlisting}

\subsubsection{D-ID3 - Prompt}
\lstinputlisting[language=Python]{prompts_and_examples/prompt__did3__text_classification.m}

\subsubsection{D-ID3 - Example Features}
\begin{lstlisting}[language=Python]
def feature(text: str) -> float:
    "Average character length of words in the text"
    words = text.split()
    if not words:
        return 0.0
    return sum(len(word) for word in words) / len(words)

def feature(text: str) -> float:
    "Calculates the proportion of text that is in passive voice"
    passive_pattern = r'\b(?:is|was|were|be|being|been) \w+\b'
    passive_count = len(re.findall(passive_pattern, text))
    return float(passive_count) / len(text.split()) if text.split() else 0.0

def feature(text: str) -> float:
    "Assesses the use of passive voice constructions in the text"
    passive_voice = re.findall(r'\b(is|are|was|were|be|being|been)\s+\w+\b', text)
    return float(len(passive_voice))

def feature(text: str) -> float:
    "Counts the number of transition words to evaluate the flow of text"
    transition_words = set(['however', 'therefore', 'moreover', 'furthermore', 'nevertheless', 'consequently'])
    words = text.lower().split()
    count = sum(1 for word in words if word in transition_words)
    return float(count)

def feature(text: str) -> float:
    "Measures the proportion of first-person pronouns in the text"
    first_person_pronouns = set(['I', 'me', 'my', 'mine', 'we', 'us', 'our', 'ours'])
    words = text.lower().split()
    count = sum(1 for word in words if word in first_person_pronouns)
    return float(count) / len(words) if words else 0.0

\end{lstlisting}

\end{document}

%% file: math_commands.tex
\usepackage{amsmath,amsfonts,bm}

\def\eqref#1{equation~\ref{#1}}
\def\1{\bm{1}}

\DeclareMathAlphabet{\mathsfit}{\encodingdefault}{\sfdefault}{m}{sl}
\SetMathAlphabet{\mathsfit}{bold}{\encodingdefault}{\sfdefault}{bx}{n}

%% file: iclr2026_conference.bbl
\begin{thebibliography}{39}
\providecommand{\natexlab}[1]{#1}
\providecommand{\url}[1]{\texttt{#1}}
\expandafter\ifx\csname urlstyle\endcsname\relax
  \providecommand{\doi}[1]{doi: #1}\else
  \providecommand{\doi}{doi: \begingroup \urlstyle{rm}\Url}\fi

\bibitem[Anthropic(2024)]{claude}
Anthropic.
\newblock Claude, 2024.
\newblock URL \url{https://www.anthropic.com}.

\bibitem[Bengio \& LeCun(2007)Bengio and LeCun]{Bengiochapter2007}
Yoshua Bengio and Yann LeCun.
\newblock Scaling learning algorithms towards {AI}.
\newblock In \emph{Large Scale Kernel Machines}. MIT Press, 2007.

\bibitem[Bommasani et~al.(2021)Bommasani, Hudson, Adeli, Altman, Arora, von
  Arx, Bernstein, Bohg, Bosselut, Brunskill,
  et~al.]{bommasani2021opportunities}
Rishi Bommasani, Drew~A Hudson, Ehsan Adeli, Russ Altman, Simran Arora, Sydney
  von Arx, Michael~S Bernstein, Jeannette Bohg, Antoine Bosselut, Emma
  Brunskill, et~al.
\newblock On the opportunities and risks of foundation models.
\newblock \emph{arXiv e-prints}, pp.\  arXiv--2108, 2021.

\bibitem[Breiman(2001)]{breiman2001random}
Leo Breiman.
\newblock Random forests.
\newblock \emph{Machine learning}, 45\penalty0 (1):\penalty0 5--32, 2001.

\bibitem[Chen et~al.(2021)Chen, Tworek, Jun, Yuan, Pinto, Kaplan, Edwards,
  Burda, Joseph, Brockman, et~al.]{chen2021evaluating}
Mark Chen, Jerry Tworek, Heewoo Jun, Qiming Yuan, Henrique Ponde De~Oliveira
  Pinto, Jared Kaplan, Harri Edwards, Yuri Burda, Nicholas Joseph, Greg
  Brockman, et~al.
\newblock Evaluating large language models trained on code.
\newblock \emph{arXiv preprint arXiv:2107.03374}, 2021.

\bibitem[Chen et~al.(2022)Chen, Ma, Wang, and Cohen]{chenprogram}
Wenhu Chen, Xueguang Ma, Xinyi Wang, and William~W Cohen.
\newblock Program of thoughts prompting: Disentangling computation from
  reasoning for numerical reasoning tasks.
\newblock \emph{Transactions on Machine Learning Research}, 2022.

\bibitem[Cheng \& Camargo(2023)Cheng and Camargo]{cheng2023chess}
Isaac Cheng and Chico Camargo.
\newblock Machine learning to study patterns in chess games.
\newblock Master's thesis, {University of Exeter}, 2023.

\bibitem[Conmy et~al.(2023)Conmy, Mavor-Parker, Lynch, Heimersheim, and
  Garriga-Alonso]{conmy2023towards}
Arthur Conmy, Augustine Mavor-Parker, Aengus Lynch, Stefan Heimersheim, and
  Adri{\`a} Garriga-Alonso.
\newblock Towards automated circuit discovery for mechanistic interpretability.
\newblock \emph{Advances in Neural Information Processing Systems},
  36:\penalty0 16318--16352, 2023.

\bibitem[Damian et~al.(2022)Damian, Lee, and Soltanolkotabi]{damian2022neural}
Alexandru Damian, Jason Lee, and Mahdi Soltanolkotabi.
\newblock Neural networks can learn representations with gradient descent.
\newblock In \emph{Conference on Learning Theory}, pp.\  5413--5452. PMLR,
  2022.

\bibitem[Dong \& Liu(2018)Dong and Liu]{dong2018feature}
Guozhu Dong and Huan Liu.
\newblock \emph{Feature engineering for machine learning and data analytics}.
\newblock CRC press, 2018.

\bibitem[Doshi-Velez \& Kim(2017)Doshi-Velez and Kim]{doshi2017towards}
Finale Doshi-Velez and Been Kim.
\newblock Towards a rigorous science of interpretable machine learning.
\newblock \emph{arXiv preprint arXiv:1702.08608}, 2017.

\bibitem[Doumbouya et~al.(2025)Doumbouya, Jurafsky, and Manning]{doumbouya2025}
Moussa Koulako~Bala Doumbouya, Dan Jurafsky, and Christopher~D. Manning.
\newblock Tversky neural networks: Psychologically plausible deep learning with
  differentiable tversky similarity.
\newblock 2025.
\newblock URL \url{https://arxiv.org/abs/2506.11035}.

\bibitem[Hewitt et~al.(2023)Hewitt, Thickstun, Manning, and
  Liang]{hewitt2023backpack}
John Hewitt, John Thickstun, Christopher~D Manning, and Percy Liang.
\newblock Backpack language models.
\newblock \emph{arXiv preprint arXiv:2305.16765}, 2023.

\bibitem[Huben et~al.(2023)Huben, Cunningham, Smith, Ewart, and
  Sharkey]{huben2023sparse}
Robert Huben, Hoagy Cunningham, Logan~Riggs Smith, Aidan Ewart, and Lee
  Sharkey.
\newblock Sparse autoencoders find highly interpretable features in language
  models.
\newblock In \emph{The Twelfth International Conference on Learning
  Representations}, 2023.

\bibitem[Hurst et~al.(2024)Hurst, Lerer, Goucher, Perelman, Ramesh, Clark,
  Ostrow, Welihinda, Hayes, Radford, et~al.]{hurst2024gpt4o}
Aaron Hurst, Adam Lerer, Adam~P Goucher, Adam Perelman, Aditya Ramesh, Aidan
  Clark, AJ~Ostrow, Akila Welihinda, Alan Hayes, Alec Radford, et~al.
\newblock Gpt-4o system card.
\newblock \emph{arXiv preprint arXiv:2410.21276}, 2024.

\bibitem[Ko et~al.(2025)Ko, Park, Lee, and Lee]{ko2025ferg}
Jeonghyun Ko, Gyeongyun Park, Donghoon Lee, and Kyunam Lee.
\newblock Ferg-llm: Feature engineering by reason generation large language
  models.
\newblock In \emph{Findings of the Association for Computational Linguistics:
  NAACL 2025}, pp.\  4211--4228, 2025.

\bibitem[Lange et~al.(2024)Lange, Tian, and Tang]{lange2024large}
Robert Lange, Yingtao Tian, and Yujin Tang.
\newblock Large language models as evolution strategies.
\newblock In \emph{Proceedings of the Genetic and Evolutionary Computation
  Conference Companion}, pp.\  579--582, 2024.

\bibitem[LeCun(1998)]{lecun1998mnist}
Yann LeCun.
\newblock The mnist database of handwritten digits.
\newblock \url{http://yann. lecun. com/exdb/mnist/}, 1998.

\bibitem[Lichess.org()]{lichess}
Lichess.org.
\newblock Lichess.
\newblock \url{https://lichess.org/about}.

\bibitem[Lundberg \& Lee(2017)Lundberg and Lee]{lundberg2017unified}
Scott~M Lundberg and Su-In Lee.
\newblock A unified approach to interpreting model predictions.
\newblock \emph{Advances in neural information processing systems}, 30, 2017.

\bibitem[Lv et~al.(2024)Lv, Xia, and Huang]{lv2024codeact}
Weijie Lv, Xuan Xia, and Sheng-Jun Huang.
\newblock Codeact: Code adaptive compute-efficient tuning framework for code
  llms.
\newblock \emph{arXiv preprint arXiv:2408.02193}, 2024.

\bibitem[Marks et~al.(2025)Marks, Rager, Michaud, Belinkov, Bau, and
  Mueller]{marks2025}
Samuel Marks, Can Rager, Eric~J. Michaud, Yonatan Belinkov, David Bau, and
  Aaron Mueller.
\newblock Sparse feature circuits: Discovering and editing interpretable causal
  graphs in language models.
\newblock 2025.
\newblock URL \url{https://arxiv.org/abs/2403.19647}.

\bibitem[Michalski et~al.(2013)Michalski, Carbonell, and
  Mitchell]{michalski2013machine}
Ryszard~Stanislaw Michalski, Jaime~Guillermo Carbonell, and Tom~M Mitchell.
\newblock \emph{Machine learning: An artificial intelligence approach}.
\newblock Springer Science \& Business Media, 2013.

\bibitem[Mu \& Andreas(2021)Mu and Andreas]{mu2021}
Jesse Mu and Jacob Andreas.
\newblock Compositional explanations of neurons.
\newblock 2021.
\newblock URL \url{https://arxiv.org/abs/2006.14032}.

\bibitem[Novikov et~al.(2025)Novikov, V{\~u}, Eisenberger, Dupont, Huang,
  Wagner, Shirobokov, Kozlovskii, Ruiz, Mehrabian,
  et~al.]{novikov2025alphaevolve}
Alexander Novikov, Ng{\^a}n V{\~u}, Marvin Eisenberger, Emilien Dupont, Po-Sen
  Huang, Adam~Zsolt Wagner, Sergey Shirobokov, Borislav Kozlovskii,
  Francisco~JR Ruiz, Abbas Mehrabian, et~al.
\newblock Alphaevolve: A coding agent for scientific and algorithmic discovery.
\newblock \emph{arXiv preprint arXiv:2506.13131}, 2025.

\bibitem[OpenAI(2025)]{openai2025gpt5}
OpenAI.
\newblock {GPT-5 System Card}.
\newblock \url{https://cdn.openai.com/gpt-5-system-card.pdf}, August 2025.

\bibitem[Pedregosa et~al.(2011)Pedregosa, Varoquaux, Gramfort, Michel, Thirion,
  Grisel, Blondel, Prettenhofer, Weiss, Dubourg, Vanderplas, Passos,
  Cournapeau, Brucher, Perrot, and Duchesnay]{scikit-learn}
F.~Pedregosa, G.~Varoquaux, A.~Gramfort, V.~Michel, B.~Thirion, O.~Grisel,
  M.~Blondel, P.~Prettenhofer, R.~Weiss, V.~Dubourg, J.~Vanderplas, A.~Passos,
  D.~Cournapeau, M.~Brucher, M.~Perrot, and E.~Duchesnay.
\newblock Scikit-learn: Machine learning in {P}ython.
\newblock \emph{Journal of Machine Learning Research}, 12:\penalty0 2825--2830,
  2011.

\bibitem[Piriyakulkij et~al.(2025)Piriyakulkij, Liang, Tang, Weller, Kryven,
  and Ellis]{piriyakulkij2025poe}
Wasu~Top Piriyakulkij, Yichao Liang, Hao Tang, Adrian Weller, Marta Kryven, and
  Kevin Ellis.
\newblock Poe-world: Compositional world modeling with products of programmatic
  experts.
\newblock \emph{Advances in Neural Information Processing Systems (to appear)},
  2025.

\bibitem[Quinlan(1986)]{quinlan1986induction}
J.~Ross Quinlan.
\newblock Induction of decision trees.
\newblock \emph{Machine learning}, 1\penalty0 (1):\penalty0 81--106, 1986.

\bibitem[Romera-Paredes et~al.(2024)Romera-Paredes, Barekatain, Novikov, Balog,
  Kumar, Dupont, Ruiz, Ellenberg, Wang, Fawzi, et~al.]{romera2024mathematical}
Bernardino Romera-Paredes, Mohammadamin Barekatain, Alexander Novikov, Matej
  Balog, M~Pawan Kumar, Emilien Dupont, Francisco~JR Ruiz, Jordan~S Ellenberg,
  Pengming Wang, Omar Fawzi, et~al.
\newblock Mathematical discoveries from program search with large language
  models.
\newblock \emph{Nature}, 625\penalty0 (7995):\penalty0 468--475, 2024.

\bibitem[Romstad et~al.(2008)Romstad, Costalba, Kiiski, Linscott, Nasu,
  Isozaki, Noda, et~al.]{stockfish2008}
Tord Romstad, Marco Costalba, Joona Kiiski, Gary Linscott, Yu~Nasu, Motohiro
  Isozaki, Hisayori Noda, et~al.
\newblock Stockfish, 2008.
\newblock URL \url{https://stockfishchess.org}.

\bibitem[Ruoss et~al.(2024)Ruoss, Del{\'e}tang, Medapati, Grau-Moya, Wenliang,
  Catt, Reid, Lewis, Veness, and Genewein]{ruoss2024amortized}
Anian Ruoss, Gr{\'e}goire Del{\'e}tang, Sourabh Medapati, Jordi Grau-Moya, Li~K
  Wenliang, Elliot Catt, John Reid, Cannada~A Lewis, Joel Veness, and Tim
  Genewein.
\newblock Amortized planning with large-scale transformers: A case study on
  chess.
\newblock \emph{Advances in Neural Information Processing Systems},
  37:\penalty0 65765--65790, 2024.

\bibitem[Sagawa et~al.(2020)Sagawa, Koh, Hashimoto, and
  Liang]{sagawa2020waterbirds}
Shiori Sagawa, Pang~Wei Koh, Tatsunori~B Hashimoto, and Percy Liang.
\newblock Distributionally robust neural networks for group shifts: On the
  importance of regularization for worst-case generalization.
\newblock \emph{International Conference on Learning Representations}, 2020.

\bibitem[Tang et~al.(2024)Tang, Key, and Ellis]{tang2024worldcoder}
Hao Tang, Darren Key, and Kevin Ellis.
\newblock Worldcoder, a model-based llm agent: Building world models by writing
  code and interacting with the environment.
\newblock \emph{Advances in Neural Information Processing Systems},
  37:\penalty0 70148--70212, 2024.

\bibitem[Verma et~al.(2024)Verma, Fleisig, Tomlin, and
  Klein]{verma2024ghostbuster}
Vivek Verma, Eve Fleisig, Nicholas Tomlin, and Dan Klein.
\newblock Ghostbuster: Detecting text ghostwritten by large language models.
\newblock In \emph{Proceedings of the 2024 Conference of the North American
  Chapter of the Association for Computational Linguistics: Human Language
  Technologies (Volume 1: Long Papers)}, pp.\  1702--1717, 2024.

\bibitem[Wang et~al.(2023)Wang, Lan, Liu, Ouyang, Qin, Lu, Chen, Zeng, and
  Yu]{oodsurvey}
Jindong Wang, Cuiling Lan, Chang Liu, Yidong Ouyang, Tao Qin, Wang Lu, Yiqiang
  Chen, Wenjun Zeng, and Philip~S. Yu.
\newblock Generalizing to unseen domains: A survey on domain generalization.
\newblock \emph{IEEE Transactions on Knowledge and Data Engineering},
  35\penalty0 (8):\penalty0 8052--8072, 2023.
\newblock \doi{10.1109/TKDE.2022.3178128}.

\bibitem[Xiao et~al.(2017)Xiao, Rasul, and Vollgraf]{xiao2017fashion}
Han Xiao, Kashif Rasul, and Roland Vollgraf.
\newblock Fashion-mnist: a novel image dataset for benchmarking machine
  learning algorithms.
\newblock \emph{arXiv preprint arXiv:1708.07747}, 2017.

\bibitem[Yao et~al.(2025)Yao, Zhang, Xi, Wang, Xu, Deng, and Chen]{yao2025}
Yunzhi Yao, Ningyu Zhang, Zekun Xi, Mengru Wang, Ziwen Xu, Shumin Deng, and
  Huajun Chen.
\newblock Knowledge circuits in pretrained transformers.
\newblock 2025.
\newblock URL \url{https://arxiv.org/abs/2405.17969}.

\bibitem[Zhang \& Liu(2024)Zhang and Liu]{zhang2024tifg}
Xinhao Zhang and Kunpeng Liu.
\newblock Tifg: Text-informed feature generation with large language models.
\newblock In \emph{2024 IEEE International Conference on Big Data (BigData)},
  pp.\  8256--8258. IEEE, 2024.

\end{thebibliography}
